\documentclass[10pt,journal,compsoc]{IEEEtran}
\usepackage{graphicx}
\usepackage{amsmath,amssymb} 
\usepackage{color}
\usepackage{float}
\usepackage{tablefootnote}
\usepackage{mathtools}
\usepackage{pbox}
\usepackage{hyperref}
\usepackage{amsmath}
\usepackage{xcolor}
\hypersetup{
    colorlinks,
    linkcolor={red},
    citecolor={green},
    urlcolor={blue}
}
\usepackage[caption=false,font=footnotesize,labelfont=sf,textfont=sf]{subfig}
\usepackage{balance}

\newcommand{\ie}{\textit{i.e., }}
\newcommand{\revision}[1]{{\color{black}{#1}}} 

\ifCLASSOPTIONcompsoc
  \usepackage[nocompress]{cite}

\else
  \usepackage{cite}
\fi

\ifCLASSINFOpdf
\else
\fi
\begin{document}
%
\title{Image Aesthetic Assessment: \\ An Experimental Survey}
%
%
%
%

\author{Yubin~Deng,
        Chen~Change~Loy,~\IEEEmembership{Member,~IEEE,}
        and~Xiaoou~Tang,~\IEEEmembership{Fellow,~IEEE}
\IEEEcompsocitemizethanks{\IEEEcompsocthanksitem Y. Deng, C. C. Loy and X. Tang are with the Department
of Information Engineering, The Chinese University of Hong Kong. \protect\\
E-mail: \{dy015, ccloy, xtang\}@ie.cuhk.edu.hk
}
}

\IEEEtitleabstractindextext{%
\begin{abstract}
This survey aims at reviewing recent computer vision techniques used in the assessment of image aesthetic quality. Image aesthetic assessment aims at computationally distinguishing high-quality photos from low-quality ones based on photographic rules, \revision{typically in the form of binary classification or quality scoring}.
A variety of approaches has been proposed in the literature trying to solve this challenging problem. In this survey, we present a systematic listing of the reviewed approaches based on visual feature types (hand-crafted features and deep features) and evaluation criteria (dataset characteristics and evaluation metrics). Main contributions and novelties of the reviewed approaches are highlighted and discussed. In addition, following the emergence of deep learning techniques, we systematically evaluate recent deep learning settings that are useful for developing a robust deep model for aesthetic scoring. Experiments are conducted using simple yet solid baselines that are competitive with the current state-of-the-arts. 
Moreover, we discuss the possibility of manipulating the aesthetics of images through computational approaches. 
We hope that our survey could serve as a comprehensive reference source for future research on the study of image aesthetic assessment.
\end{abstract}

\begin{IEEEkeywords}
Aesthetic quality classification, image aesthetics manipulations
\end{IEEEkeywords}}

\maketitle

\IEEEdisplaynontitleabstractindextext

%
\IEEEpeerreviewmaketitle

\IEEEraisesectionheading{\section{Introduction}\label{sec:introduction}}

\begin{figure*}
\centering
\includegraphics[width=\linewidth]{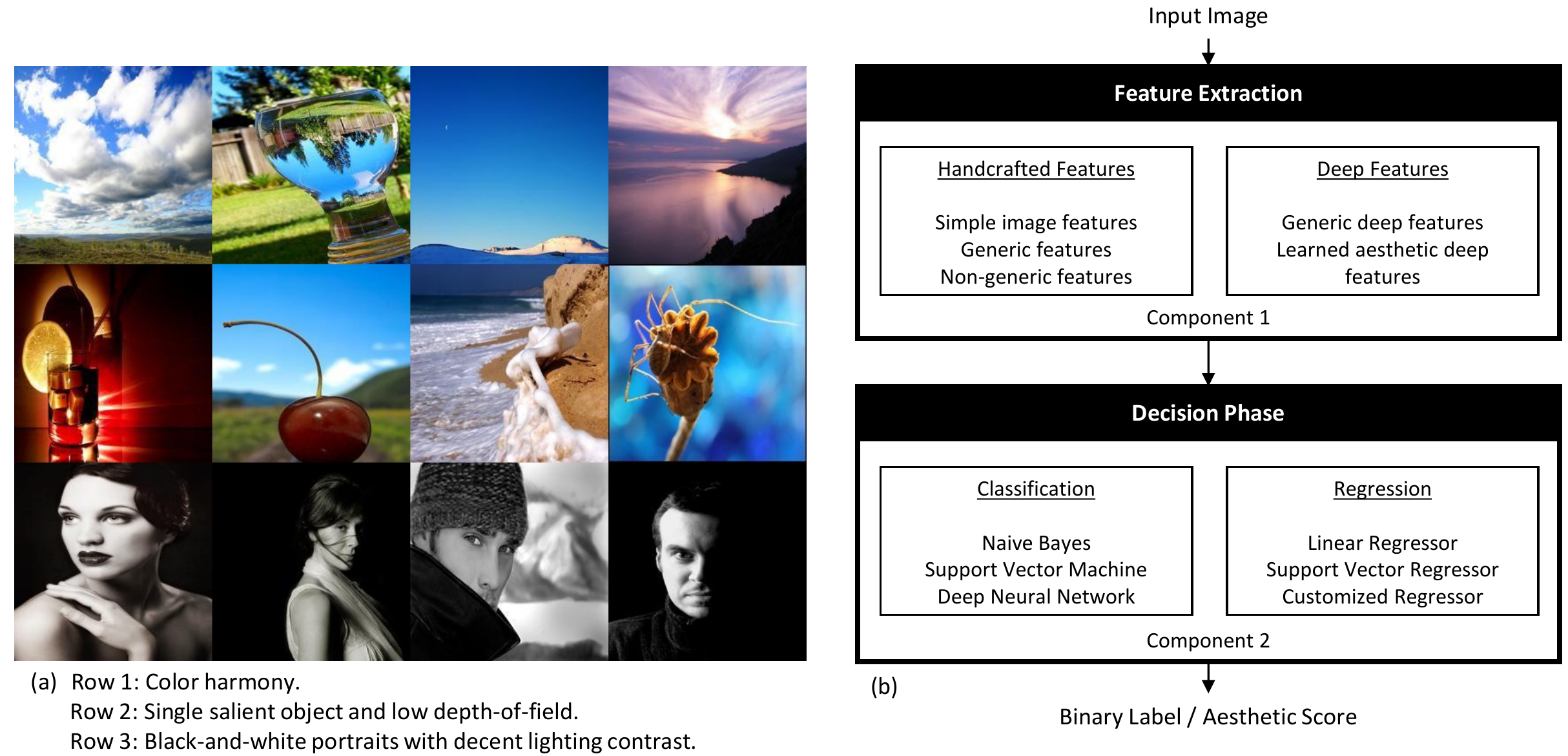}
\vskip -0.4cm
\caption{(a) High-quality images following well-established photographic rules. (b) A typical flow of image aesthetic assessment systems.}
\label{fig:typicalFlow}
\end{figure*}

\IEEEPARstart{T}{he}
aesthetic quality of an image is judged by commonly established photographic rules, which can be affected by numerous factors including the different usages of lighting~\cite{freeman2007complete}, contrast~\cite{itten2005design}, and image composition~\cite{london2005photography} (see Fig.~\ref{fig:typicalFlow}a). 
These human judgments given in an aesthetic evaluation setting are the results of human aesthetic experience, \ie the interaction between emotional-valuation, sensory-motor, and meaning-knowledge neural systems, as demonstrated in a systematic neuroscience study by Chatterjee et al.~\cite{chatterjee2016neuroscience}. 
From the beginning of psychological aesthetics studies by Fechner~\cite{fechner1876vorschule} to modern neuroaesthetics, researchers argue that there is a certain connection between human aesthetic experience and the sensation caused by visual stimuli regardless of source, culture, and experience~\cite{zeki2013clive}, which is supported by activations in specific regions of the visual cortex~\cite{ishizu2013brain,brown2011naturalizing,barrett2007experience,barrett2006structure}. For example, human's general reward circuitry produces pleasure when people look at beautiful objects~\cite{kuhn2012neural}, and the subsequent aesthetic judgment consists of the appraisal of the valence of such perceived objects~\cite{brown2011naturalizing,barrett2007experience,barrett2006structure,leder2004model}. These activations in the visual cortex can be attributed to the processing of various early, intermediate and late visual features of the stimuli including orientation, shape, color grouping and categorization~\cite{chatterjee2004prospects,greenlee2008functional,wandell2009visual,zeki1999inner}. Artists intentionally incorporate such features to facilitate desired perceptual and emotional effects in viewers, forming a set of guidelines as they create artworks to induce desired responses in the nervous systems of perceivers~\cite{zeki1999inner,cavanagh2005artist}.  And modern day photographers now resort to certain well-established photographic rules~\cite{ang2002digital,freeman2007photographer} when they capture images as well, in order to make their work appealing to a large group of audiences.

As the volume of visual data available online grows at an exponential rate, the capability of automatically distinguishing high-quality images from low-quality ones gain increasing demands in real-world image searching and retrieving applications. In image search engines, it is expected that the systems \color{black}{will }\color{black} return professional photographs instead of random snapshots when a particular keyword is entered. For example, when a user enters ``mountain scenery'', he/she will expect to see colorful, pleasing mountain \color{black}{views }\color{black} or well-captured mountain peaks instead of gray or blurry mountain snapshots.

The design of these intelligent systems can potentially be facilitated by insights from neuroscience studies, which show that human aesthetic experience is a kind of information processing that includes five stages: perception, implicit memory integration, explicit classification of content and style, cognitive mastering and evaluation, and ultimately produces aesthetic judgment and aesthetic emotion~\cite{chatterjee2004prospects,leder2004model}. However, it is non-trivial to computationally model this process. Challenges in the task of judging the quality of an image include (i) computationally modeling the intertwined photographic rules, (ii) knowing the aesthetical differences in images \color{black}{from }\color{black} different image genres (e.g., close-shot object, profile, scenery, night scenes), (iii) knowing the type of techniques used in photo capturing (e.g., HDR, black-and-white, depth-of-field), and (iv) obtaining a large amount of human-annotated data for robust testing. 

To address these challenges, computer vision researchers typically cast this problem as a classification or regression problem. Early studies started with distinguishing typical snapshots from professional \color{black}{photographs }\color{black} by trying to model the well-established photographic rules using low-level features~\cite{tong2004classification,datta2006studying,liu2010optimizing}. These systems typically involve a training set and a testing set consisting of high-quality images and low-quality \color{black}{ones}\color{black}. The system robustness is judged by the model performance on the testing set using a specified metric such as accuracy. 
These rule-based approaches are intuitive as they try to explicitly model the criteria that humans use in evaluating the aesthetic quality of an image. However, more recent studies~\cite{lu2014rapid,lu2015deep,kong2016photo,mai2016composition} have shown that using a data-driven approach is more effective, as the amount of training data available grows from a couple of hundreds of images to millions of images. Besides, transfer learning from source tasks with sufficient amount of data to a target task with relatively fewer training data is also proven feasible, with many successful attempts showing promising results by deep learning methods~\cite{lecun2015deep} with network fine-tune, where image aesthetics are implicitly learned in a data-driven manner.  

As summarized in Fig.~\ref{fig:typicalFlow}b, the majority of aforementioned computer vision approaches for image aesthetic assessment can be categorized based on image representations (e.g., handcrafted features and learned features) and classifiers/regressors training (e.g., Support Vector Machine (SVM) and neural network learning \color{black}{approaches}\color{black}). To our best knowledge, there does not exist up-to-date survey that covers the state-of-the-art methodologies involved in image aesthetic assessment.  The last review was published in 2011 by Joshi et al.~\cite{joshi2011aesthetics}, and no deep learning based methods were covered. Some reviews on image quality assessment have been published~\cite{ebrahimi2015subjective,george2014survey}. In those lines of effort, image quality metrics regarding the differences between a noise-tempered sample and the original high-quality image have been proposed, including but not limited to mean squared error (MSE), structural similarity index (SSIM)~\cite{wang2004image} and visual information fidelity (VIF)~\cite{sheikh2005visual}. 
Nevertheless, their main focus is on distinguishing noisy images from clean images in terms of a different quality measure, rather than artistic/photographic aesthetics.


In this article, we wish to contribute a thorough overview of the field of image aesthetic assessment; meanwhile, we will also cover the basics of deep learning methodologies. Specifically, as different datasets exist and evaluation criteria vary in the image aesthetics literature, we do not aim at directly comparing the system performance of all reviewed work; instead, in the survey we point out their main contributions and novelties in model designs, and give potential insights for future directions in this field of study. 
In addition, following the recent emergence of deep learning techniques and the effectiveness of the data-driven approach in learning better image representations, we systematically evaluate different techniques that could facilitate the learning of a robust deep classifier for aesthetic scoring. Our study covers topics including data preparation, fine-tune strategies, and multi-column deep architectures, which we believe to be useful for researchers working in this domain. \color{black}{In particular, we summarize useful insights on how to alleviate the potential problem of data distribution bias in a binary classification setting and show the effectiveness of rejecting false positive predictions using our proposed convolution neural network (CNN) baselines,  as revealed by the balanced accuracy metric}\color{black}.
Moreover, we also review the most commonly used publicly available image aesthetic assessment datasets for this problem and draw connections between image aesthetic assessment and image aesthetic manipulation, including image enhancement, computational photography and automatic image cropping. We hope that this survey can serve as a  comprehensive reference source and inspire future research in understanding image aesthetics and its many potential applications.

\subsection{Organization}
The rest of this paper is organized as follows. We first give a review of deep neural network basics and recap the objective image quality metrics in section~\ref{sec:background}.
Then in Section~\ref{sec:pipeline}, we explain the typical pipeline used by the majority of the reviewed work on this problem and highlight the most concerned design component. We review existing datasets in Section~\ref{sec:datasets}. We present a review on conventional methods based on handcrafted features in Section~\ref{sec:handcraft_feature} and deep features in Section~\ref{sec:deep_feature}. Evaluation criteria and existing results are discussed in Section~\ref{sec:evaluation_criteria}.
In Section~\ref{sec:evaluation}, we systematically analyze various deep learning settings using a baseline model that is competitive with the state-of-the-arts. 
In Section~\ref{sec:image_aesthetic_manipulation}, we draw a connection between aesthetic assessment and aesthetic manipulation, with a focus on aesthetic-based image cropping. Finally, we conclude with a discussion of the current state of research and give some recommendations for future directions on this field of study. 

\section{Background}
\label{sec:background}
\subsection{Deep Neural Network}
\label{sec:deep_neural_network}
Deep neural network belongs to the family of deep learning methods that are tasked to learn feature representation in a data-driven approach. \revision{While shallow models (e.g., SVM, boosting) have shown success in earlier literatures concerning relatively smaller amounts of data, they require highly-engineered feature designs in solving machine learning problems}.
Common architectures in deep neural networks consist of a stack of parameterized individual modules that we call ``layers'', such as convolution layer and fully-connected layer. The architecture design of stacking layers on top of layers is inspired by the hierarchy in human visual cortex ventral pathway, offering different levels of abstraction for the learned representation in each layer.
Information propagation among layers in feed-forward deep neural networks typically follows a sequential manner. A ``forward'' operation  $F(\cdot)$ is defined respectively in each layer to propagate the input $\boldsymbol{x}$ it receives and produces an output $\boldsymbol{y}$ to the next layer. 
For example, the forward operation in a fully-connected layer with learnable weights $\mathbf{W}$ can be written as:
\begin{equation}
	y = F(\boldsymbol{x}) = \mathbf{W}\boldsymbol{x} = \sum w_{ij}\cdot x_{i}
\end{equation}
This is typically followed by a non-linear function, such as sigmoid 
\begin{equation}
z = \frac{1}{1+\exp(-y)}
\end{equation} 
or the rectified linear unit (ReLU) $z = \max(0, y)$, which acts as the activation function and produces the net activation output $z$.

To learn the weights $\mathbf{W}$ in a data-driven manner, we need to have the feedback information that reports the current performance of the network. Essentially, we are trying to tune the knobs $\mathbf{W}$ in order to achieve a learning objective. For example, given an objective $t$ for the input $\boldsymbol{x}$, we want to minimize the squared error between the net output $z$ and $t$ by defining a loss function L:
\begin{equation}
L = \frac{1}{2}|| z - t ||^{2}
\end{equation}
To propagate this feedback information to the weights, we define the ``backward'' operation for each layer using the gradient back-propagation~\cite{lecun1989handwritten}. We hope to get the direction $\Delta \mathbf{W}$ to update the weights $\mathbf{W}$ in order to better suit the training objective (\ie to minimize $L$): $\mathbf{W} \leftarrow \mathbf{W} - \eta \Delta \mathbf{W}$,  where $\eta$ is the learning rate. In our example, $\Delta \mathbf{W}$ can be easily derived based on the chain rule:
\begin{equation}
\begin{aligned}
\Delta \mathbf{W} &= \frac{\partial L}{\partial \mathbf{W}} \\
&= \frac{\partial L}{\partial z}\frac{\partial z}{\partial y}\frac{\partial y}{\partial \mathbf{W}} \\
&= (z-t)\cdot \frac{\exp(-y)}{(\exp(-y)+1)^2} \cdot \boldsymbol{x}
\end{aligned}
\end{equation}
In practice, researchers resort to batch stochastic gradient descent (SGD) or more advanced learning procedures that compute more stable gradients as averaged from a batch of training examples $\{(\boldsymbol{x}_i, t_i) | \boldsymbol{x}_{i} \in X\}$ in order to train deeper and deeper neural networks with continually increasing amounts of layers. We refer readers to~\cite{lecun2015deep} for an in-depth overview of more deep learning methodologies.

\subsection{Image Quality Metrics}
\label{sec:metrics}

\begin{figure}
\centering
\includegraphics[width=\linewidth]{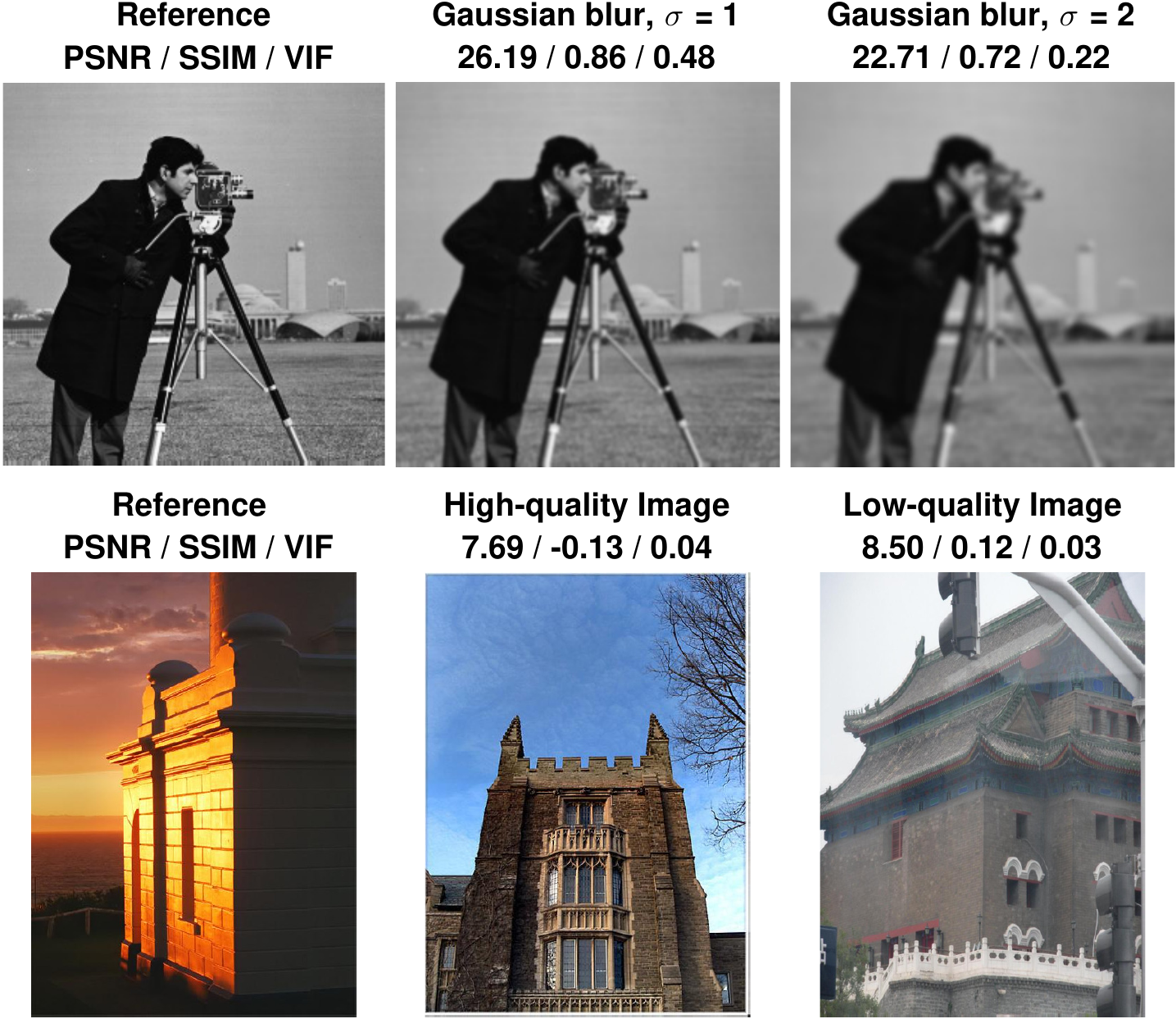}
\vskip -0.3cm
\caption{Quality measurement by Peak Signal-to-Noise Ratio (PSNR), Structural Similarity Index (SSIM)~\cite{wang2004image} and Visual Information Fidelity (VIF)~\cite{sheikh2005visual} (higher is better, typically measured against a referencing groundtruth high-quality image). Although these are good indicators for measuring the quality of images in image restoration applications as in Row 1, they do not reflect human perceived aesthetic values as shown by the measurements for the building images in Row 2.}
\label{fig:quality_metric}
\end{figure}

\revision{Image quality metrics are defined in an attempt to quantitatively measure the objective quality of an image. This is typically used in image restoration applications (super-resolution~\cite{dong2016image}, de-blur~\cite{shan2008high} and de-artifacts~\cite{dong2015compression}) where we have a default high-quality reference image for comparison.
However, these quality metrics are not designed to measure the subjective nature of human perceived aesthetic quality (see examples in Fig.~\ref{fig:quality_metric}). Directly applying these objective quality metrics to our concerned domain of image aesthetic assessment may produce misleading results, as can be seen from the measured values in the second row of Fig.~\ref{fig:quality_metric}. Developing more robust metrics has gained increasing interests in the research community in an attempt to assess the more subjective image aesthetic quality.}

\section{A Typical Pipeline}
\label{sec:pipeline}

Most existing image quality assessment methods take a supervised learning approach. A typical pipeline assumes a set of training data $\left \{  \boldsymbol{x}_{i}, y_{i} \right \}_{i \in [1, N] }$, from which a function $f: g(X) \rightarrow Y$ is learned, where $g(\boldsymbol{x}_{i})$ denotes the feature representation of image $\boldsymbol{x}_{i}$. The label $\mathbf{y}_{i}$ is either $\{0, 1\}$ for binary classification (when $f$ is a classifier) or a continuous score range for regression (when $f$ is a regressor). 
Following this formulation, a pipeline can be broken into two main components as shown in Fig.~\ref{fig:typicalFlow}b, \ie a feature extraction component and a decision component.

\subsection{Feature Extraction}
The first component of an image aesthetics assessment system aims at extracting robust feature representations describing the aesthetic aspect of an image. Such features are assumed to model the photographic/artistic aspect of images in order to distinguish images of different qualities. Numerous efforts have been seen in designing features that are robust enough for the intertwined aesthetic rules. The majority of feature types can be classified into handcrafted features and deep features. Conventional approaches
~\cite{datta2006studying,ke2006design,tong2004classification,sun2009photo,you2009perceptual,luo2008photo,nishiyama2011aesthetic,lo2012statistic,bhattacharya2010framework,dhar2011high,tang2013content,yeh2012relative,marchesotti2011assessing,marchesotti2013learning,murray2012ava} typically adopt handcrafted features to computationally model the photographic rules (lighting, contrast), global image layout (rule-of-thirds) and typical objects (human profiles, animals, plants) in images. In more recent work, generic deep features~\cite{dong2015photo,lv2016learning} and learned deep features~\cite{peng2016toward,wang2016multi,tian2015query,lu2015deep,lu2014rapid,lu2015improvedrapid,wang2016brain,zhang2016describing,kong2016photo,kao2016hierarchical,kao2016visual} exhibit stronger representation power for this task.  

\subsection {Decision Phase}
The second component of an image aesthetics assessment system provides the ability to perform classification or regression for the given aesthetic task. Na\"{i}ve Bayes classifier, SVM, boosting and deep classifier are typically used for binary classification of high-quality and low-quality images, whereas regressors like support vector regressor are used in ranking or scoring images based on their aesthetic quality.

\begin{figure*}
\centering
\includegraphics[width=\linewidth]{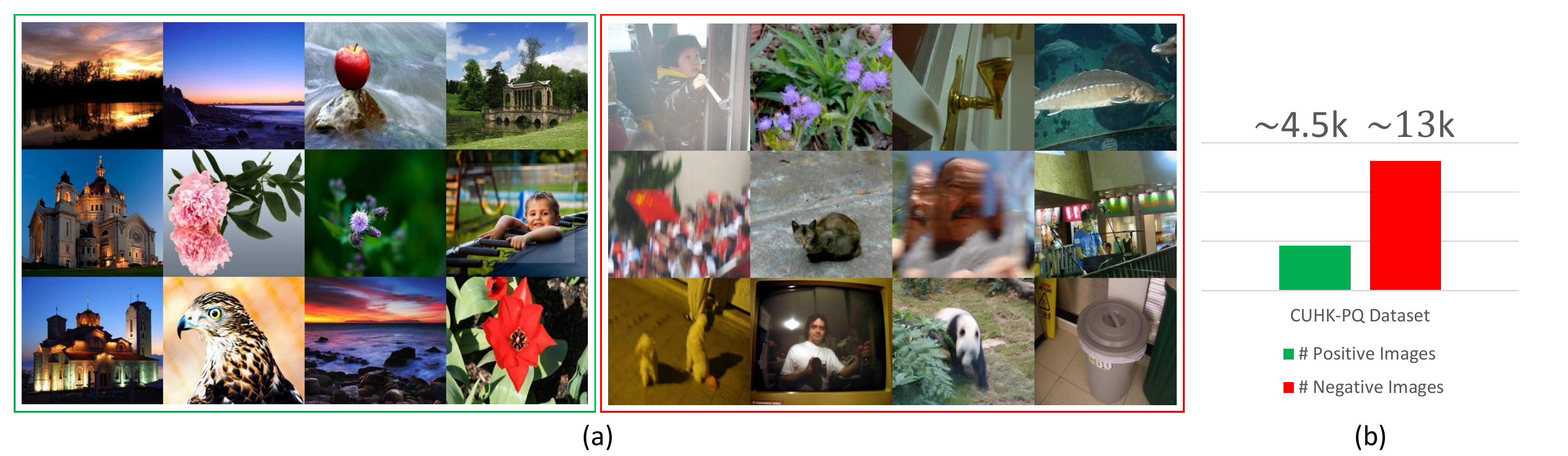}
\vskip -0.5cm
\caption{(a) Sample images in the CUHK-PQ dataset. Distinctive differences can be visually observed between the high-quality images (grouped in green) and low-quality ones (grouped in red). (b) Number of images in CUHK-PQ dataset.}
\label{fig:teaseImg-CUHK}
\end{figure*}

\begin{figure*}
\centering
\includegraphics[width=\linewidth]{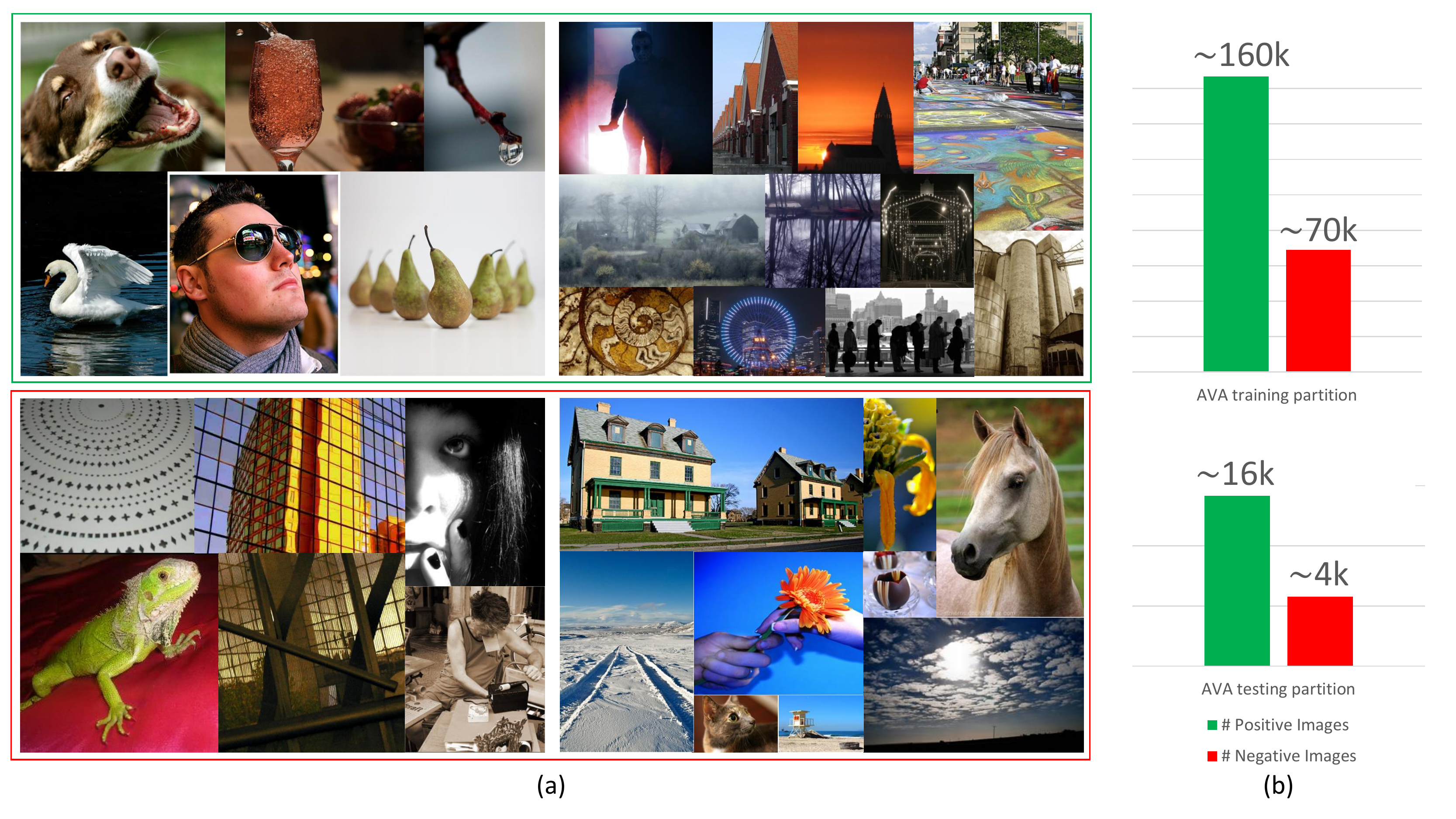}
\vskip -0.5cm
\caption{(a) Sample images in the AVA dataset. Top: images labeled with mean score $>$ 5, grouped in green. Bottom: images labeled with mean score $<$ 5, grouped in red. The image groups on the right are ambiguous ones having a somewhat neutral scoring around 5. (b) Number of images in AVA dataset.}
\label{fig:teaseImg-AVA}
\vspace{-0.1cm}
\end{figure*}

\section{Datasets}
\label{sec:datasets}

The assessment of image aesthetic quality assumes a standard training set and testing set containing high-quality image examples and low-quality ones, as mentioned above. Judging the groundtruth aesthetic quality of a given image is, however, a subjective task. As such, it is inherently challenging to obtain a large amount of such annotated data. Most of the earlier papers~\cite{datta2006studying,sun2009photo,you2009perceptual} on image aesthetic assessment collect a small amount of private image data. These datasets typically contain from a few hundred to a few thousand images with binary labels or aesthetic scoring for each image. Yet, such datasets where the model performance is evaluated are not publicly available. Much research effort has later been made to contribute publicly available image aesthetic datasets of larger scales for more standardized evaluation of model performance. In the following, we introduce those datasets that are most frequently used in performance benchmarking for image aesthetic assessment.

\textbf{The Photo.Net dataset} and \textbf{The DPChallenge dataset} are introduced in~\cite{datta2008algorithmic,joshi2011aesthetics}. These two datasets can be considered as the earliest attempt to construct large-scale image database for image aesthetic assessment. The Photo.Net dataset contains 20,278 images with at least 10 score ratings per image. The rating ranges from 0 to 7 with 7 assigned to the most aesthetically pleasing photos. \color{black}Typically, images uploaded to Photo.net are rated as somewhat pleasing, with the peak of the global mean score skewing to the right in the distribution~\cite{joshi2011aesthetics}\color{black}. The more challenging DPChallenge dataset contains diverse rating. The DPChallenge dataset contains 16,509 images in total, and has been later replaced by the AVA dataset, where a significantly larger amount of images derived from DPChallenge.com are collected and annotated.

\textbf{The CUHK-PhotoQuality (CUHK-PQ) dataset} is introduced in~\cite{luo2011content,tang2013content}. It contains 17,690 images collected from DPChallenge.com and amateur photographers. All images are given binary aesthetic label and grouped into 7 scene categories, \ie ``animal'', ``plant'', ``static'', ``architecture'', ``landscape'', ``human'', and ``night''. The standard training and testing set from this dataset are random partitions of 50-50 split or a 5-fold cross validation partition, where the overall ratio of the total number of positive examples and that of the negative examples is around $1:3$. Sample images are shown in Fig.~\ref{fig:teaseImg-CUHK}.

\textbf{The Aesthetic Visual Analysis (AVA) dataset}~\cite{murray2012ava} contains $\sim250k$ images in total. These images are obtained from DPChallenge.com and labeled by aesthetic scores. Specifically, each image receives $78\sim549$ votes of score ranging from 1 to 10. The average score of an image is commonly taken to be its groundtruth label. As such, it contains more challenging examples as images lie within the center score range could be ambiguous in their aesthetic aspect (Fig.~\ref{fig:teaseImg-AVA}a). For the task of binary aesthetic quality classification, images with an average score higher than threshold $5+\sigma$ are treated as positive examples, and images with an average score lower than $5-\sigma$ are treated as negative ones. Additionally, the AVA dataset contains 14 style attributes and more than 60 category attributes for a subset of images. There are two typical training and testing splits from this dataset, \ie (i) a large-scale standardized partition with $\sim230k$ training images and $\sim20k$ testing images \revision{ using a hard threshold $\sigma=0 $} (ii) and an easier partition modeling that of CUHK-PQ by taking those images whose score ranking is at top $10\%$ and bottom $10\%$, resulting in $\sim25k$ images for training and $\sim25k$ images for testing. The ratio of the total number of positive examples and that of the negative examples is around $12:5$.

Apart from these two standard benchmarks, more recent research also introduce new datasets that take into consideration the data-balancing issue. \textbf{The Image Aesthetic Dataset (IAD)} introduced in~\cite{lu2015improvedrapid} contains 1.5 million images derived from DPChallenge and PHOTO.NET. Similar to AVA, images in the IAD dataset are scored by annotators. Positive examples are selected from those images with a mean score larger than a threshold. All IAD images are used for model training, and the model performance is evaluated on AVA in~\cite{lu2015improvedrapid}. The ratio of the number of positive examples and that is the negative examples is around $1.07:1$. \textbf{The Aesthetic and Attributes DataBase (AADB)}~\cite{kong2016photo} also contains a balanced distribution of professional and consumer photos, with a total of $10,000$ images. Eleven aesthetic attributes and annotators' ID is provided. A standard partition with 8,500 images for training, 500 images for validation, and 1,000 images for testing is proposed~\cite{kong2016photo}.

The trend to creating datasets of even larger volumes and higher diversity is essential for boosting the research progress in this field of study. To date, the AVA dataset serves as a canonical benchmark for performance evaluation of image aesthetic assessment as it is the first large-scale dataset with detailed annotation.
Still, the distribution of positive examples and negative ones in the dataset also play a role in the effectiveness of trained models, as false positive predictions are as harmful as having low recall rate in image retrieval/searching applications. In the following, we review major attempts in the literature to build systems for the challenging task of image aesthetic assessment.

\section{Conventional Approaches with Handcrafted Features}
\label{sec:handcraft_feature}

The conventional option for image quality assessment is to hand design good feature extractors, which requires a considerable amount of engineering skill and domain expertise. Below we review a variety of approaches that exploit hand-engineered features.

\subsection{Simple Image Features}
Global features are first explored by researchers to model the aesthetic aspect of images. The work by Datta et al.~\cite{datta2006studying} and Ke et al.~\cite{ke2006design} are among the first to cast aesthetic understanding of images into a binary classification problem. Datta et al.~\cite{datta2006studying} combine low-level features and high-level features that are typically used for image retrieval and train an SVM classifier for binary classification of images aesthetic quality. Ke et al.~\cite{ke2006design} propose global edge distribution, color distribution, hue count and low-level contrast and brightness indicators to represent an image, then they train a Na\"{i}ve Bayes classifier based on such features. An even earlier attempt by Tong et al.~\cite{tong2004classification} adopt boosting to combine global low-level simple features (blurriness, contrast, colorfulness, and saliency) in order to classify professional photograph and ordinary snapshots. All these pioneering works present the very first attempts to computationally modeling the global aesthetic aspect of images using handcrafted features. Even in a recent work, Ayd{\i}n et al.~\cite{aydin2015automated} construct image aesthetic attributes by sharpness, depth, clarity, tone, and colorfulness. An overall aesthetics rating score is heuristically computed based on these five attributes. Improving upon these global features, later studies adopt global saliency to estimate aesthetic attention distribution. Sun et al.~\cite{sun2009photo} make use of global saliency map to estimate visual attention distribution to describe an image, and they train a regressor to output the quality score of an image based on the rate of focused attention region in the saliency map. You et al.~\cite{you2009perceptual} derive similar attention features based on global saliency map and incorporate temporal activity feature for video quality assessment. 

Regional image features~\cite{luo2008photo,nishiyama2011aesthetic,lo2012statistic} later prove to be effective in complementing the global features. Luo et al.~\cite{luo2008photo} extract regional clarity contrast, lighting, simplicity, composition geometry, and color harmony features based on the subject region of an image.
Wong et al.~\cite{wong2009saliency} compute exposure, sharpness and texture features on salient regions and global image, as well as 
features depicting the subject-background relationship of an image.
%
Nishiyama et al~\cite{nishiyama2011aesthetic} extract bags-of-color-patterns from local image regions with a grid-sampling technique. While~\cite{luo2008photo,wong2009saliency,nishiyama2011aesthetic} adopt the SVM classifier, Lo et al.~\cite{lo2012statistic} build a statistic modeling system with coupled spatial relations after extracting color and texture feature from images, where a likelihood evaluation is used for aesthetic quality prediction. These methods focus on modeling image aesthetics from local image regions that are potentially most attracted to humans.

\begin{figure}
\centering
\includegraphics[width=\linewidth]{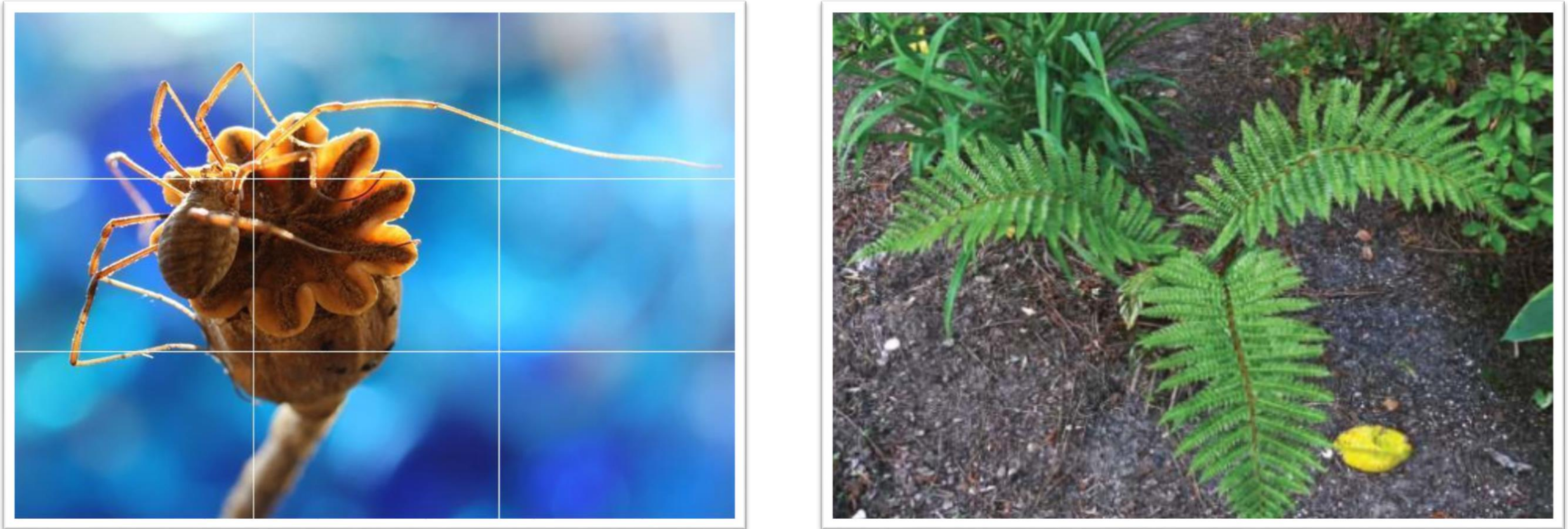}
\vskip -0.3cm
\caption{Left: Image composition with low depth-of-field, single salient object, and rule-of-thirds. Right: Image of low aesthetic quality.}
\label{fig:rule-of-thirds}
\end{figure}

\subsection{Image Composition Features}
Image composition in a photograph typically relates to the presence and position of a salient object. Rule-of-thirds, low depth-of-field, and opposing colors are the common techniques for composing a good image where the salient object is made outstanding (see Fig.~\ref{fig:rule-of-thirds}). To model such aesthetic aspect, Bhattacharya et al.~\cite{bhattacharya2010framework,bhattacharya2011holistic} propose compositional features using relative foreground position and visual weight ratio to model the relations between foreground objects and the background scene, then a support vector regressor is trained. Wu et al.~\cite{wu2010good} propose the use of Gabor filter responses to estimate the position of the main object in images, and extract low-level HSV-color features from global and central image regions. These features are fed to a soft-SVM classifier with sigmoidal softening in order to distinguish images of ambiguous quality. Dhar et al.~\cite{dhar2011high} cast high-level features into describable attributes of composition, content and sky illumination and combine low-level features to train an SVM classifier. Lo et al.~\cite{lo2012assessment} propose the combination of layout composition, edge composition features with HSV color palette, HSV counts and global features (textures, blur, dark channel, contrasts). SVM is used as the classifier. 

The representative work by Tang et al.~\cite{tang2013content} give a comprehensive analysis of the fusion of global features and regional features. Specifically, image composition is estimated by global hue composition and scene composition, and multiple types of regional features extracted from subject areas are proposed, such as dark channel feature, clarity contrast, lighting contrast, composition geometry of the subject region, spatial complexity and human-based features. An SVM classifier is trained on each of the features for comparison and the final model performance is substantially enhanced by combining all the proposed features. It is shown that regional features can effectively complement global features in modeling the images aesthetics.

A more recent approach by image composition features is proposed by Zhang et al.~\cite{zhang2014fusion} where image descriptors that characterize local and global structural aesthetics from multiple visual channels are designed. Spatial structure of the image local regions are modeled using graphlets, and they are connected based on atomic region adjacency. To describe such atomic regions, visual features from multiple visual channels (such as color moment, HOG, saliency histogram) are used. The global spatial layout of the photo are also embedded into graphlets using a Grassmann manifold. The importances of the two kinds of graphlet descriptors are dynamically adjusted, capturing the spatial composition of an image from multiple visual channels. The final aesthetic prediction of an image is generated by a probabilistic model using the post-embedding graphlets. 

\subsection{\revision{General-Purpose Features}}

Yeh et al.~\cite{yeh2012relative} make use of SIFT descriptors and propose relative features by matching a query photo to photos in a gallery group. General-purpose imagery features like Bag-of-Visual-Words (BOV)~\cite{csurka2004visual} and Fisher Vector  (FV)~\cite{perronnin2007fisher} are explored in~\cite{marchesotti2011assessing,marchesotti2013learning,murray2012ava}. Specifically, SIFT and color descriptors are used as the local descriptors upon which a Gaussian Mixture Model (GMM) is trained. The statistics up to the second order of this GMM distribution is then encoded using BOV or FV. Spatial pyramid is also adopted and the per-region encoded FV's are concatenated as the final image representation. These methods~\cite{marchesotti2011assessing,marchesotti2013learning,murray2012ava} represent the attempt to implicitly modeling photographic rules by encoding them in generic content based features, which is competitive or even outperforms the simple handcrafted features.

\subsection{\revision{Task-Specific Features}}
\revision{Task-specific features} refer to features in image aesthetic assessment that are optimized for a specific category of photos, which can be efficient when the use-case or task scenario is fixed or known beforehand. Explicit information (such as human face characteristics, geometry tag, scene information, intrinsic character component properties) is exploited based on the different task nature.

Li et al.~\cite{li2010aesthetic} propose a regression model that targets only consumer photos with faces. Face-related social features (such as face expression features, face pose features, relative face position features) and perceptual features (face distribution symmetry, face composition, pose consistency) are specifically designed for measuring the quality of images with faces, and it is shown in~\cite{li2010aesthetic} that they complement with conventional handcrafted features (brightness contrast, color correlation, clarity contrast and background color simplicity) for this task. Support vector regression is used to produce aesthetic scores for images.

Lienhard et al.~\cite{lienhard2015low} study particular face features for evaluating the aesthetic quality of headshot images. To design features for face/headshots, the input image is divided into sub-regions (eyes region, mouth region, global face region and entire image region). Low-level features (sharpness, illumination, contrast, dark channel, hue and saturation in the HSV color space) are computed from each region. These pixel-level features assume the human perception while viewing a face image, hence can reasonably model the headshot images. SVM with Gaussian kernel is used as the classifier. 

Su et al.~\cite{su2011scenic} propose bag-of-aesthetics-preserving features for scenic/landscape photographs. Specifically, an image is decomposed into $n \times n$ spatial grids, then low-level features in HSV-color space, as well as LBP, HOG and saliency features are extracted from each patch. The final feature is generated by a predefined patch-wise operation to exploit the landscape composition geometry. AdaBoost is used as the classifier. These features aim at modeling only the landscape images and may be limited in their representation power in general image aesthetic assessment.

Yin et al.~\cite{yin2012assessing} build a scene-dependent aesthetic model by incorporating the geographic location information with GIST descriptors and spatial layout of saliency features for scene aesthetic classification (such as bridges, mountains and beaches). SVM is used as the classifier. The geographic location information is used to link a target scene image to relevant photos taken within the same geo-context, then these relevant photos are used as the training partition to the SVM. Their proposed model requires input images with geographic tags and is also limited to the scenic photos. For scene images without geo-context information, SVM trained with images from the same scene category is used.  

Sun et al.~\cite{sun2015aesthetic} design a set low-level features for aesthetic evaluation of Chinese calligraphy. They target the handwritten Chinese character in a plain-white background; hence conventional color information is not useful in this task. Global shape features, extracted based on standard calligraphic rules, are introduced to represent a character. In particular, they consider alignment and stability, distribution of white space, stroke gaps as well as a set of component layout features while modeling the aesthetics of handwritten characters. A back-propagation neural network is trained as the regressor to produce an aesthetic score for each given input.

\section{Deep Learning Approaches}
\label{sec:deep_feature}

The powerful feature representation learned from a large amount of data has shown an ever-increased performance on the tasks of recognition, localization, retrieval, and tracking, surpassing the capability of conventional handcrafted features~\cite{krizhevsky2012imagenet}. Since the work by Krizhevsky et al.~\cite{krizhevsky2012imagenet} where convolutional neural networks (CNN) is adopted for image classification, mass amount of interest is spiked in learning robust image representations by deep learning approaches. 
Recent works in the literature of image aesthetic assessment using deep learning approaches to learn image representations can be broken down into two major schemes, (i) adopting generic deep features learned from other tasks and training a new classifier for image aesthetic assessment and (ii) learning aesthetic deep features and classifier directly from image aesthetics data. 

\subsection{Generic Deep Features}
A straightforward approach to employ deep learning approaches is to adopt generic deep features learned from other tasks and train a new classifier on the aesthetic classification task. Dong et al.~\cite{dong2015photo} propose to adopt the generic features from penultimate layer output of AlexNet~\cite{krizhevsky2012imagenet} with spatial pyramid pooling. Specifically, the $4096 \ (\textit{fc7}) \times 6 \ (Spatial Pyramid) = 24576$ dimensional feature is extracted as the generic representation for images, then an SVM classifier is trained for binary aesthetic classification. Lv et al.~\cite{lv2016learning} also adopt the normalized 4096-dim \textit{fc7} output of AlexNet~\cite{krizhevsky2012imagenet} as feature presentation. They propose to learn the relative ordering relationship of images of different aesthetic quality. They use $SVM^{rank}$~\cite{joachims2006training} to train a ranking model for image pairs of $\{I_{highQuality}, I_{lowQuality}\}$. 

\begin{figure}[t]
\centering
\includegraphics[width=\linewidth]{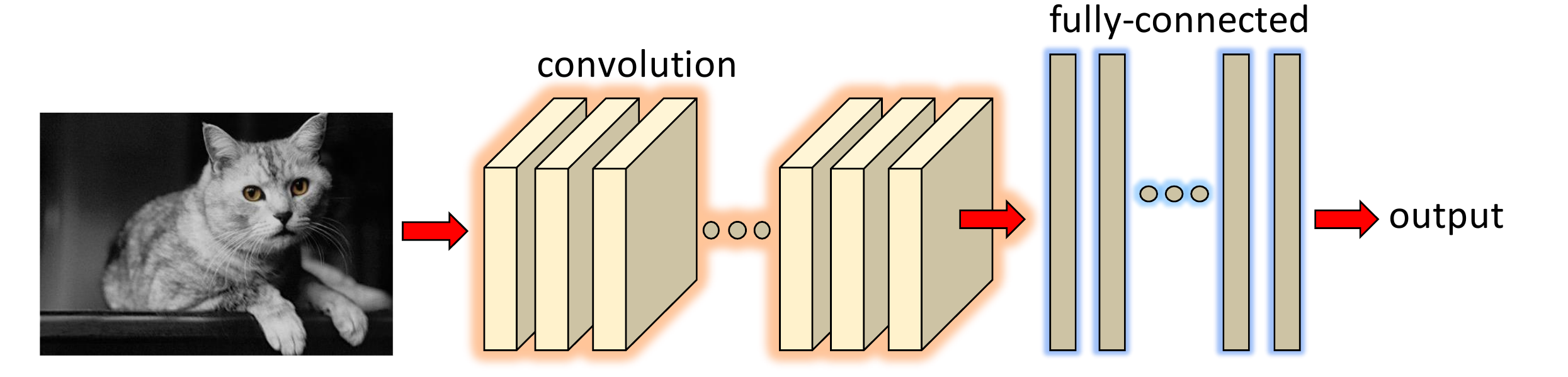}
\vskip -0.3cm
\caption{The architecture of typical single-column CNNs.}
\label{fig:network-single}
\end{figure}

\begin{figure}[t]
\centering
\includegraphics[width=\linewidth]{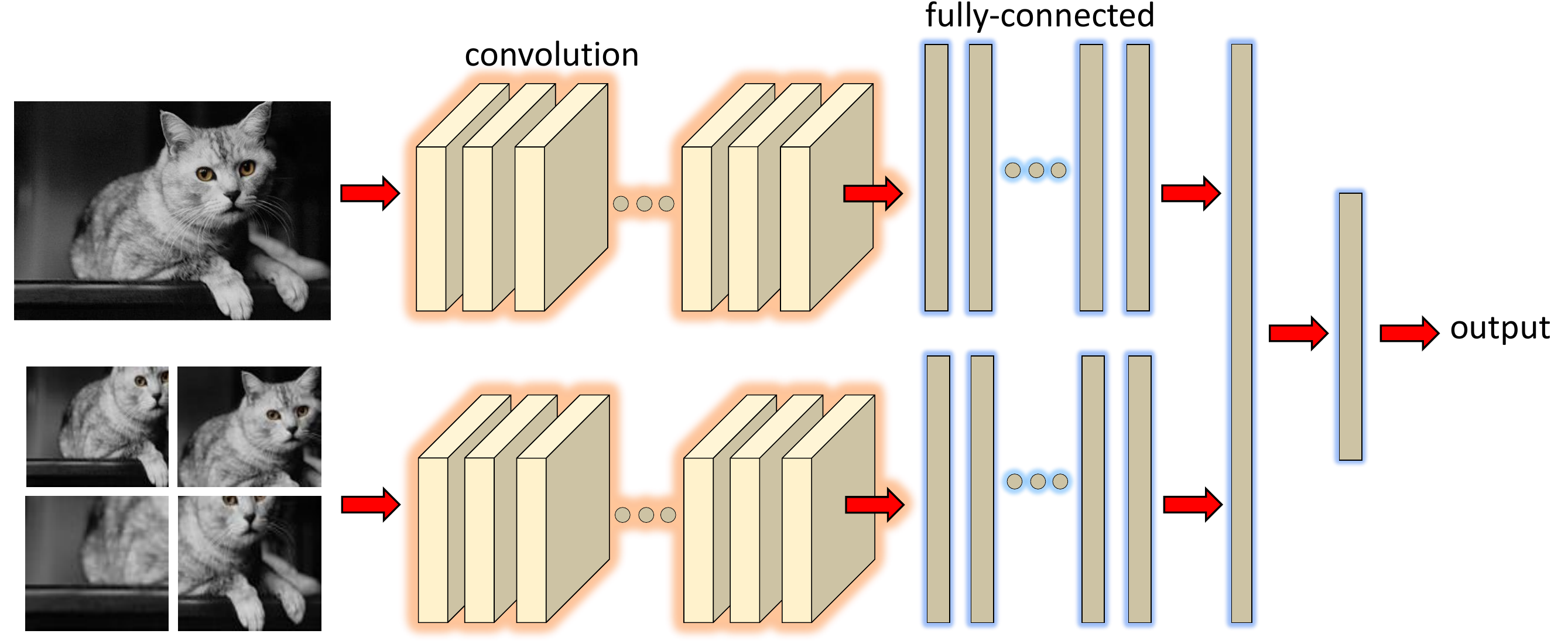}
\vskip -0.3cm
\caption{Typical multi-column CNN: a two-column architecture is shown as an example.}
\label{fig:network-double}
\end{figure}

\subsection{Learned Aesthetic Deep Features}
\textbf{Features learned with single-column CNNs (Fig.~\ref{fig:network-single}):}
Peng et al.~\cite{peng2016toward} propose to train CNNs of AlexNet-like architecture for 8 different abstract tasks (emotion classification, artist classification, artistic style classification, aesthetic classification, fashion style classification, architectural style classification, memorability prediction, and interestingness prediction). In particular, the last layer of the CNN for aesthetic classification is modified to output 2-dim softmax probabilities. This CNN is trained from scratch using aesthetic data, and the penultimate layer ($fc7$) output is used as the feature representation. To further analyze the effectiveness of the features learned from other tasks, Peng et al. analyze different pre-training and fine-tune strategies and evaluate the performance of different combinations of the concatenated $fc7$ features from the 8 CNNs.

Wang et al.~\cite{wang2016multi} propose a CNN that is modified from the AlexNet architecture. Specifically, the $conv_5$ layer of AlexNet is replaced by a group of 7 convolutional layers (with respect to different scene categories), which are stacked in a parallel manner with mean pooling before feeding to the fully-connected layers, \ie $\{conv_5^{1-animal},conv_5^{2-architecture},conv_5^{3-human},\\conv_5^{4-landscape},conv_5^{5-night},conv_5^{6-plant},conv_5^{7-static}\}$. The fully connected layers \textit{fc6} and \textit{fc7} are modified to output 512 feature maps instead of 4096 for more efficient parameters learning. The 1000-class softmax output is changed to 2-class softmax (\textit{fc8}) for binary classification. The advantage of this CNN using such a group of 7 parallel convolutional layers is to exploit the aesthetic aspects in each of the 7 scene categories. During pre-training, a set of images belonging to 1 of the scene categories is used for each one of the $conv_5^{i} (i \in \{1, ..., 7\})$  layers, then the weights learned through this stage is transferred back to the $conv_5^{i}$ in the proposed parallel architecture, with the weights from $conv_1$ to $conv_4$ reused from AlexNet the weights in the fully-connected layer randomly re-initialized. Subsequently, the CNN is further fine-tuned end-to-end. Upon convergence, the network produces a strong response in the $conv_5^{i}$ layer feature map when the input image is of category $i \in \{1, ..., 7\}$. This shows the potential in exploiting image category information when learning the aesthetic presentation. 

Tian et al.~\cite{tian2015query} train a CNN with 4 convolution layers and 2 fully-connected layers to learn aesthetic features from data. The output size of the 2 fully-connected layers is set to 16 instead of 4096 as in AlexNet. The authors propose that such a 16-dim representation is sufficient to model only the top $10\%$ and bottom $10\%$ of the aesthetic data, which is relatively easy to classify compared to the full data. Based on this efficient feature representation learned from CNN, the authors propose a query-dependent aesthetic model as the classifier. Specifically, for each query image, a query-dependent training set is retrieved based on predefined rules (visual similarity, image tags association, or the combination of both). Subsequently, an SVM is trained on this retrieved training set. It shows that the features learned from aesthetic data outperform the generic deep features learned in the ImageNet task.  

The DMA-net is proposed in~\cite{lu2015deep} where information from multiple image patches are extracted by a single-column CNN that contains 4 convolution layers and 3 fully-connected layers, with the last layer outputting a softmax probability. Each randomly sampled image patch is fed into this CNN. To combine multiple feature output from the sampled patches of one input image, a statistical aggregation structure is designed to aggregate the features from the orderless sampled image patches by multiple poolings (min, max, median and averaging). An alternative aggregation structure is also designed based on sorting. The final feature representation effectively encodes the image based on regional image information.

\noindent
\textbf{Features learned from Multi-column CNNs (Fig.~\ref{fig:network-double}):}
The \textit{RAPID} model by Lu et al.~\cite{lu2014rapid,lu2015improvedrapid} can be considered to be the first attempt in training convolutional neural networks with aesthetic data. They use an AlexNet-like architecture where the last fully-connected layer is set to output 2-dim probability for aesthetic binary classification. Both global image and local image patch are considered in their network input design, and the best model is obtained by stacking a global-column and a local-column CNN to form a double-column CNN (DCNN), where the feature representation (penultimate layers \textit{fc7} output) from each column is concatenated before the \textit{fc8} layer (classification layer). Standard Stochastic Gradient Descent (SGD) is used to train the network with softmax loss. Moreover, they further boost the performance of the network by incorporating image style information using a style-column or semantic-column CNN. Then the style-column CNN is used as the third input column, forming a three-column CNN with style/semantic information (SDCNN). Such multi-column CNN exploits the data from both global and local aspect of images.

Mai et al.~\cite{mai2016composition} propose stacking 5-columns of VGG-based networks using an adaptive spatial pooling layer. The adaptive spatial pooling layer is designed to allow arbitrary sized image as input; specifically, it pools a fixed-length output given different receptive field sizes after the last convolution layer. By varying the kernel size of the adaptive pooling layer, each sub-network effectively encodes multi-scale image information. Moreover, to potentially exploit the aesthetic aspect of different image categories, a scene categorization CNN outputs a scene category posterior for each of the input image, then a final scene-aware aggregation layer processes such aesthetic features (category posterior \& multi-scale VGG features) and outputs the final classification label. The design of this multi-column network has the advantage to exploit the multi-scale compositions of an image in each sub-column by adaptive pooling, yet the multi-scale VGG features may contain redundant or overlapping information, and could potentially lead to network overfitting.

Wang et al.~\cite{wang2016brain} propose a multi-column CNN model called \textit{BDN} that share similar structures with \textit{RAPID}. In \textit{RAPID}, a style attribute prediction CNN is trained to predict 14 styles attributes for input images. This attribute-CNN is treated as one additional CNN column, which is then added to the parallel input pathways of a global image column and a local patch column. In \textit{BDN}, 14 different style CNNs are pre-trained and they are parallel cascaded and used as the input to a final CNN for rating distribution prediction, where the aesthetic quality score of an image is subsequently inferred. The \textit{BDN} model can be considered as an extended version of \textit{RAPID} that exploits each of the aesthetic attributes using learned CNN features, hence enlarging the parameter space and learning capability of the overall network.

Zhang et al.~\cite{zhang2016describing} propose a two-column CNN for learning aesthetic feature representation. The first column ($CNN_1$) takes image patches as input and the second column ($CNN_2$) takes a global image as input. Instead of randomly sampling image patches given an input image, a weakly-supervised learning algorithm is used to project a set of $D$ textual attributes learned from image tags to highly-responsive image regions. Such image regions in images are then fed to the input of $CNN_1$. This $CNN_1$ contains 4 convolution layers and one fully-connected layers ($fc_5$) at the bottom, then a parallel group of $D$ output branches ($fc_6^i, i \in \{1,2, ..., D\}$) modeling each one of the $D$ textual attributes are connected on top. The size of the feature maps of each of the $fc_6^i$ is of 128-dimensional. A similar $CNN_2$ takes a globally warped image as input, producing one more 128-dim feature vector from $fc_6$. Hence, the final concatenated feature learned in this manner is of $128 \times (D+1)$-dimensional. A probabilistic model containing 4 layers is trained for aesthetic quality classification.

\begin{table*}[t]
\centering
\caption{Overview of typical evaluation criteria.} 
\vskip -0.2cm
\label{Criteria-table} 
\begin{tabular}{lll} 
Method                              & Formula  		& Remarks               		  \\  \hline
Overall accuracy      & $\frac{ TP +  TN}{P + N} $		&  \pbox{20cm}{Accounting for the proportion of correctly classified samples. \\$TP$: true positive, $TN$: true negative, $P$: total positive, $N$: total negative }          \\ \\ 
Balanced accuracy  			    & $\frac{1}{2}\frac{ TP}{P} + \frac{1}{2}\frac{ TN}{N} $  & \pbox{20cm}{Averaging precision and true negative prediction for imbalanced distribution.\\ $TP$: true positive, $TN$: true negative, $P$: total positive, $N$: total negative}      \\ \\
Precision-recall curve                   & $p=\frac{ TP}{TP+FP}, r=\frac{ TP}{TP+FN}$  & \pbox{20cm}{Measuring the relationship between precision and recall.\\$TP$: true positive, $TN$: true negative, $FP$: false positive, $FN$: false negative}        \\ \\
Euclidean distance & $\sqrt{\sum_{i} (Y_{i} - \widehat{Y}_{i})^{2} }$ & \pbox{20cm}{Measuring the difference between the groundtruth score and aesthetic ratings. \\$Y$: ground truth score, $\widehat{Y}$: predicted score}           \\ \\
Correlation ranking  & $\frac{cov(rg_{X}, rg_{Y})}{\sigma_{rg_{X}} \sigma_{rg_{Y}}}$ & \pbox{20cm}{Measuring the statistical dependence between the ranking of aesthetic\\  prediction and groundtruth. \\$rg_{X}, rg_{Y}$: rank variables, $\sigma$: standard deviation, $cov$: covariance            }   \\ \\ 
ROC curve            & $tpr=\frac{ TP}{TP+FN}, fpr=\frac{ FP}{FP+TN}$  & \pbox{20cm}{Measuring model performance change by $tpr$ (true positive rate) and $fpr$ \\(false positive rate) when the binary discrimination threshold is varied. \\$TP$: true positive, $TN$: true negative, $FP$: false positive, $FN$: false negative  }  \\ \\
Mean average precision                              & $\frac{1}{n}\sum_{i}^{n} (precision(i) \times \Delta recall(i))$         & \pbox{20cm}{The averaged AP values, based on precision and recall. \\$precision(i)$ is calculated among the first $i$ predictions, $\Delta recall(i)$: change in recall} \\ \hline         
\end{tabular}
\end{table*}

\begin{table*}[t]
\centering
\caption{Methods evaluated on the CUHK-PQ dataset.}
\vskip -0.2cm
\label{CUHK-table}
\begin{tabular}{lllll}
Method                                                                                   & Dataset  & Metric               & Result  & Training-Testing Remarks    \\  \hline
Su et al. (2011)~\cite{su2011scenic}      & CUHK-PQ. & Overall accuracy             & 92.06\% & 1000 training, 3000 testing \\
Marchesotti et al. (2011)~\cite{marchesotti2011assessing}  & CUHK-PQ  & Overall accuracy             & 89.90\% & 50-50 split                 \\
Zhang et al. (2014)~\cite{zhang2014fusion} & CUHK-PQ  & Accuracy             & 90.31\% & 50-50 split, 12000 subset                 \\
Dong et al. (2015)~\cite{dong2015photo}                  & CUHK-PQ  & Overall accuracy             & 91.93\% & 50-50 split                 \\
Tian et al. (2015)~\cite{tian2015query}  & CUHK-PQ  & Overall accuracy             & 91.94\% & 50-50 split                 \\
Zhang et al. (2016)~\cite{zhang2016describing}   & CUHK-PQ  & Overall accuracy             & 88.79\% & 50-50 split, 12000 subset   \\
Wang et al. (2016)~\cite{wang2016multi}               & CUHK-PQ  & Overall accuracy             & 92.59\% & 4:1:1 partition             \\
Lo et al. (2012)~\cite{lo2012assessment}                                 & CUHK-PQ  & Area under ROC curve & 0.93    & 50-50 split                 \\
Tang et al. (2013)~\cite{tang2013content}                                        & CUHK-PQ  & Area under ROC curve & 0.9209  & 50-50 split                 \\
Lv et al. (2016)~\cite{lv2016learning}                 & CUHK-PQ  & Mean AP              & 0.879   & 50-50 split   \\
\hline              
\end{tabular}
\end{table*}

\begin{table*}[t]
\centering
\caption{Methods evaluated on the AVA dataset.}
\vskip -0.2cm
\label{AVA-other-table}
\begin{tabular}{lllll}
Method                                                                                           & Dataset     & Metric    & Result                     & Training-Testing Remarks \\ \hline
Marchesotti et al. (2013)~\cite{marchesotti2013learning}                                                & AVA & ROC curve & tp-rate: 0.7, fp-rate: 0.4 & standard partition       \\ 

AVA handcrafted features (2012)~\cite{murray2012ava}  & AVA & Overall accuracy & 68.00\% & standard partition       \\
SPP (2015)~\cite{lu2015deep}    & AVA & Overall accuracy & 72.85\% & standard partition       \\
RAPID - full method (2014)~\cite{lu2014rapid}   & AVA & Overall accuracy & 74.46\% & standard partition       \\
Peng et al. (2016)~\cite{peng2016toward}  & AVA & Overall accuracy & 74.50\% & standard partition       \\
Kao et al. (2016)~\cite{kao2016hierarchical}  & AVA & Overall accuracy & 74.51\% & standard partition       \\
RAPID - improved version (2015)~\cite{lu2015improvedrapid}  & AVA & Overall accuracy & 75.42\% & standard partition       \\
DMA net (2015)~\cite{lu2015deep}   & AVA & Overall accuracy & 75.41\% & standard partition       \\
Kao et al. (2016)~\cite{kao2016visual} & AVA & Overall accuracy & 76.15\% & standard partition       \\
Wang et al. (2016)~\cite{wang2016multi}  & AVA & Overall accuracy & 76.94\% & standard partition       \\
Kong et al. (2016)~\cite{kong2016photo}  & AVA & Overall accuracy & 77.33\% & standard partition       \\
BDN (2016)~\cite{wang2016brain}  & AVA & Overall accuracy & 78.08\% & standard partition       \\ 

Zhang et al. (2014)~\cite{zhang2014fusion}                     & AVA         & Overall accuracy  & 83.24\%                      & 10\%-subset, 12.5k*2       \\
Dong et al. (2015)~\cite{dong2015photo}                         & AVA         & Overall accuracy  & 83.52\%                      & 10\%-subset, 19k*2       \\
Tian et al. (2016)~\cite{tian2015query}         & AVA         & Overall accuracy  & 80.38\%                    & 10\%-subset, 20k*2       \\
Wang et al. (2016)~\cite{wang2016multi}                      & AVA         & Overall accuracy  & 84.88\%                    & 10\% subset, 25k*2       \\
Lv et al. (2016)~\cite{lv2016learning}                            & AVA         & Mean AP   & 0.611                      & 10\%-subset, 20k*2       \\
\hline 
\end{tabular}
\end{table*}

Kong et al.~\cite{kong2016photo} propose to learn aesthetic features assisted by the pair-wise ranking of image pairs as well as the image attribute and content information. Specifically, a Siamese architecture which takes image pairs as input is adopted, where the two base networks of the Siamese architecture adopt the AlexNet  configurations (the 1000-class classification layer \textit{fc8} from the AlexNet is removed). In the first stage, the base network is pre-trained by fine-tune from aesthetic data using Euclidean Loss regression layer instead of softmax classification layer. After that, the Siamese network ranks the loss for every sampled image pairs. Upon convergence, the fine-tuned base-net is used as a preliminary feature extractor. In the second stage, an attribute prediction branch is added to the base-net to predict image attributes information, then the base-net continues to be fine-tuned using a multi-task manner by combining the rating regression Euclidean loss, attribute classification loss and ranking loss. In the third stage, yet another content classification branch is added to the base-net in order to predict a predefined set of category labels. Upon convergence, the softmax output of the content category prediction is used as a weighting vector for weighting the scores produced by each of the feature branch (aesthetic branch, attribute branch, and content branch). In the final stage, the base-net with all the added output branches is fine-tuned jointly with content classification branch frozen. Effectively, such aesthetic features are learned by considering both the attribute and category content information, and the final network produces image scores for each given image.

\begin{figure}[t]
\centering
\includegraphics[width=\linewidth]{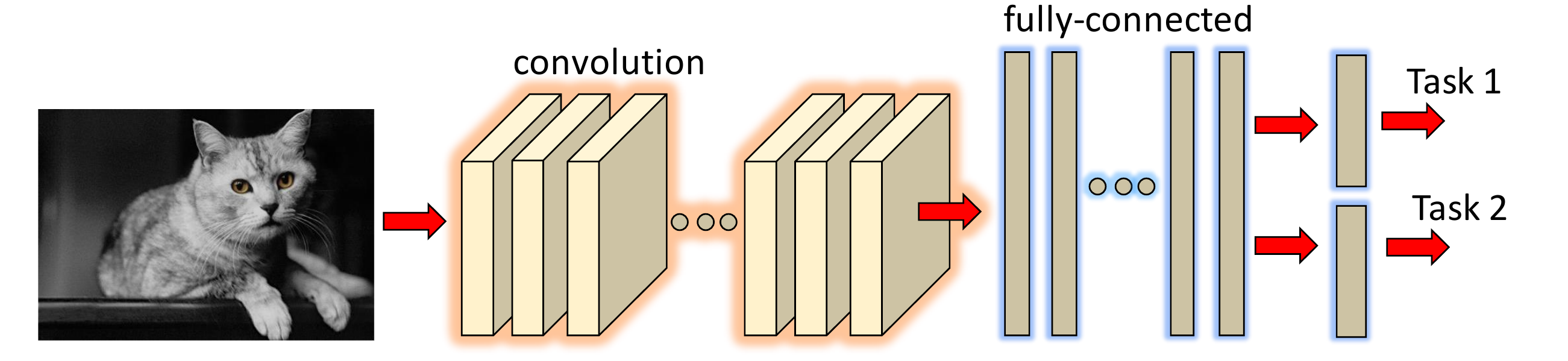}
\vskip -0.3cm
\caption{A typical multi-task CNN consists of a main task (Task 1) and multiple auxiliary tasks (only one Task 2 is shown here).}
\label{fig:network-multi}
\end{figure}

\noindent
\textbf{Features learned with Multi-Task CNNs (Fig.~\ref{fig:network-multi}):}
Kao et al.~\cite{kao2016hierarchical} propose three category-specific CNN architectures, one for object, one for scene and one for texture. The scene CNN takes warped global image as input. It has 5 convolution layers and three fully-connected layer with the last fully-connected layer producing a 2-dim softmax classification; the object CNN takes both the warped global image and the detected salient region as input. It is a 2-column CNN combining global composition and salient information; the texture CNN takes 16 randomly cropped patches as input. Category information is predicted using a 3-class SVM classifier before feeding images to a category-specific CNN. To alleviate the use of the SVM classifier, an alternative architecture with warped global image as input is trained with a multi-task approach, where the main task is aesthetic classification and the auxiliary task is scene category classification. 

Kao et al.~\cite{kao2016visual} propose to learn image aesthetics in a multi-task manner. Specifically, AlexNet is used as the base network. Then the 1000-class \textit{fc8} layer is replaced by a 2-class aesthetic prediction layer and a 29-class semantic prediction layer. The loss balance between the aesthetic prediction task and the semantic prediction task is determined empirically.
Moreover, another branch containing two fully-connected layers for aesthetic prediction is added to the second convolution layer ($conv_2$ of AlexNet).  By linking an added gradients flow from the aesthetic task directly to convolutional layers, one expects to learn better low-level convolutional features. This strategy shares a similar spirit to deeply supervised net~\cite{lee2015deeply}.

\section{Evaluation Criteria and Existing Results}
\label{sec:evaluation_criteria}

Different metrics for performance evaluation of image aesthetic assessment models are used across the literature: 
classification accuracy~\cite{datta2006studying,tong2004classification,luo2008photo,wong2009saliency,wu2010good,bhattacharya2010framework,bhattacharya2011holistic,
marchesotti2011assessing,murray2012ava,yin2012assessing,dong2015photo,peng2016toward,lienhard2015low,wang2016multi,tian2015query,lu2015deep,lu2014rapid,lu2015improvedrapid,
wang2016brain,zhang2016describing,kong2016photo,kao2016hierarchical,kao2016visual}
\revision{reports the proportion of correctly classified results;}
precision-and-recall curve~\cite{ke2006design,luo2008photo,nishiyama2011aesthetic,dhar2011high,lo2012assessment}
\revision{considers the degree of relevance of the retrieved items and the retrieval rate of relevant items, which is also widely adopted in image search or retrieval applications;}
Euclidean distance or residual sum-of-squares error between the groundtruth score and aesthetic ratings~\cite{sun2009photo,li2010aesthetic,lienhard2015low,sun2015aesthetic}
and correlation ranking~\cite{you2009perceptual,yeh2012relative,kong2016photo} 
\revision{are used for performance evaluation in score regression frameworks;}
ROC curve~\cite{lo2012statistic,lo2012assessment,marchesotti2013learning,lienhard2015low,su2011scenic}
and area under the curve~\cite{lo2012assessment,luo2011content,tang2013content}
\revision{concerns the performance of binary classifier when the discrimination threshold gets varied;}
mean average precision~\cite{lv2016learning,lu2015deep,lu2014rapid,lu2015improvedrapid} 
\revision{
is the average precision across multiple queries, which is usually used to summarize the PR-curve for the given set of samples.
}
These are among the typical metrics for evaluating model effectiveness on image aesthetic assessment (see Table~\ref{Criteria-table} for summary). 
Subjective evaluation by conducting human surveys is also seen in~\cite{aydin2015automated} where human evaluators are asked to give subjective aesthetic attribute ratings. 

We found that it is not feasible to directly compare all methods as different datasets and evaluation criteria are used across the literature. 
To this end, we try to summarize respectively the released results reported on the two standard datasets, namely CUHK-PQ (Table~\ref{CUHK-table}) and AVA datasets (Table~\ref{AVA-other-table}), and present the results on other datasets in Table~\ref{Private-table}.
%
%
%
To date, the AVA dataset (standard partition) is considered to be the most challenging dataset by the majority of the reviewed work.

\begin{table*}[t]
\centering
\caption{Methods evaluated on other datasets.}
\vskip -0.2cm
\label{Private-table}
\begin{tabular}{llll}
Method                                                                                       & Dataset                                      & Metric                        & Result                     \\ \hline
Tong et al. (2004)~\cite{tong2004classification}                                      & 29540-image private set                      & Overall accuracy                      & 95.10\%                    \\
Datta et al. (2006)~\cite{datta2006studying}                                          & 3581-image private set                       & Overall accuracy                      & 75\%                       \\
Sun et al. (2009)~\cite{sun2009photo}                                                 & 600-image private set                        & Euclidean distance            & 3.5135                     \\
Wong et al. (2009)~\cite{wong2009saliency}                                            & 3161-image private set & Overall accuracy                      & 79\%                       \\
Bhattacharya. (2010, 2011) ~\cite{bhattacharya2010framework,bhattacharya2011holistic} & $\sim$650-image private set                  & Overall accuracy                      & 86\%                       \\
Li et al. (2010)~\cite{li2010aesthetic}                                               & 500-image private set                        & Residual sum-of-squares error & 2.38                       \\
Wu et al. (2010)~\cite{wu2010good}                                                    & 10800-image private set from Flickr          & Overall accuracy      & $\sim$83\%                 \\
Dhar et al. (2011)~\cite{dhar2011high}                                                & 16000-image private set from DPChallenge     & PR-curve                      & -                          \\
Nishiyama et al. (2011)~\cite{nishiyama2011aesthetic}                     & 12k-image private set from DPChallenge       & Overall accuracy                      & 77.60\%                    \\
Lo et al. (2012)~\cite{lo2012statistic}                                               & 4k-image private set                         & ROC curve                     & tp-rate: 0.6, fp-rate: 0.3 \\
Yeh et al. (2012)~\cite{yeh2012relative}                                              & 309-image private set                        & Kendall’s Tau-b measure       & 0.2812                     \\
Aydin et al. (2015)~\cite{aydin2015automated}                                         & 955-image subset from DPChallenge.com        & Human survey         & -                        \\
                                                                                             &                                              &                               &                            \\
Yin et al. (2012)~\cite{yin2012assessing}                                             & 13k-image private set from Flickr            & Overall accuracy                      & 81\%                       \\
Lienhard et al. (2015)~\cite{lienhard2015low}                                         & Human Face Scores 250-image dataset          & Overall accuracy  & 86.50\%                    \\
Sun et al. (2015)~\cite{sun2015aesthetic}                                             & 1000-image Chinese Handwriting               & Euclidean distance            & -                          \\
Kong et al. (2016)~\cite{kong2016photo}                                               & AADB dataset                                 & Spearman ranking              & 0.6782                     \\
Zhang et al. (2016)~\cite{zhang2016describing}                                        & PNE                                          & Overall accuracy                      & 86.22\%  \\
\hline                  
\end{tabular}
\end{table*}

The overall accuracy metric appears to be the most popular metric. It can be written as 
\begin{equation}
\label{eq:overall_acc}
\textrm{Overall accuracy} = 
\frac{ TP +  TN}{P +  N}.
\end{equation}
\color{black}This metric alone could be biased and far from ideal as a Na\"{i}ve predictor that predicts all examples as positive would already reach about $(14k+0)/(14k+6k) =70\%$ classification accuracy. 
To complement such metric when evaluating models on imbalanced testing sets, an alternative balanced accuracy metric~\cite{brodersen2010balanced} can be adopted:
\begin{equation}
\label{eq:balanced_acc}
\textrm{Balanced accuracy} = 
\frac{1}{2}(\frac{TP}{P}) + \frac{1}{2}(\frac{ TN }{ N}). 
\end{equation}
Balanced accuracy equally considers the classification performance on different classes~\cite{brodersen2010balanced,huang2016learning}. 
\revision{
While overall accuracy in Eq.~\eqref{eq:overall_acc} offers an intuitive sense of correctness by reporting the proportion of correctly classified samples, balanced accuracy in Eq.~\eqref{eq:balanced_acc} combines the prevalence-independent statistics of sensitivity and specificity.
} 
A low balanced accuracy will be observed if a given classifier tends to predict only the dominant class. For the Na\"{i}ve predictor mentioned above, the balanced accuracy would give a proper number indication of $0.5\times(14k/14k) + 0.5\times(0k/6k) =50\%$ performance on AVA.

In this regard, in the following sections where we discuss our findings on a proposed strong baseline, we report both overall classification accuracy and balanced accuracy in order to get a more reasonable measure of baseline performance. 

\section{Experiments on Deep Learning Settings}
\label{sec:evaluation}

It is evident from Table~\ref{AVA-other-table} that deep learning based approaches dominate the performance of image aesthetic assessment.
The effectiveness of learned deep features in this task has motivated us to take a step back to consider how in a de facto manner that CNN works in understanding the aesthetic quality of an image. 
It is worth noting that training a robust deep aesthetic scoring model is non-trivial, and often, we found that `the devil is in the details'. 
To this end, we design a set of systematic experiments based on a baseline 1-column CNN and a 2-column CNN, and evaluate different settings from mini-batch formation to complex multi-column architecture. Results are reported on the widely used AVA dataset.

We observe that by carefully training the CNN architecture, the 2-column CNN baseline reaches comparable or even better performance with the state-of-the-arts and the 1-column CNN baseline acquires the strong capability to suppress false positive predictions while having competitive classification accuracy. 
We wish that the experimental results could facilitate designs of future deep learning models for image aesthetic assessment.

\subsection{Formulation and the Base CNN Structure}

The supervised learning process of CNNs involves a set of training data $\left \{  \boldsymbol{x}_{i}, y_{i} \right \}_{i \in [1, N] }$, from which a nonlinear mapping function $f: X \rightarrow Y$ is learned through backpropagation~\cite{le1990handwritten}. Here, $\boldsymbol{x}_{i}$ is the input to the CNN and $y_{i} \in \mathbb{T}$ is its corresponding ground truth label. For the task of binary classification, $y_{i} \in \left \{0, 1\right \}$ is the aesthetic label corresponding to image $\boldsymbol{x}_{i}$. 
\color{black} The convolutional operations in such a CNN can be expressed as 
\begin{equation}
F_{k}(X)=\text{max}(\mathbf{w}_{k}*F_{k-1}(X)+\mathbf{b}_{k}, 0),  
k \in \left \{1, 2, ..., D\right \}  \; 
\end{equation}
where $F_{0}(X) = X$ is the network input and $D$ is the depth of the convolutional layers. The operator '$*$' denotes the convolution operation. The operations in the $D'$ fully-connected layers can be formulated in a similar manner. To learn the $(D+D')$ network weights $\mathbf{W}$ using the standard backpropagation with stochastic gradient descent, we adopt the cross-entropy classification loss, which is formulated as 
\begin{equation}
\label{eq:cross_entropy_loss}
\begin{multlined}
L(\mathbf{W}) = -\frac{1}{n}\sum_{i =1}^{n} \sum_{t}\{ t~\text{log}~p(\widehat{y}_{i}=t|\boldsymbol{x}_{i};\mathbf{W})  \\ \;\;\;\;\;\;\;\;\;\;\;\;+(1-t)\text{log}(1-p(\widehat{y}_{i}=t|\boldsymbol{x}_{i};\mathbf{W}))  +\phi(\mathbf{W})  \}  \\
\end{multlined}
\end{equation}
\begin{equation}
p(\widehat{y}_{i}=t|\boldsymbol{x}_{i};\mathbf{w_{t}}) = \frac{\text{exp}(\mathbf{w}_{t}^\mathsf{T}\boldsymbol{\boldsymbol{x}_{i}})}{\sum_{t' \in \mathbb{T}} \text{exp}(\mathbf{w}_{t'}^\mathsf{T}\boldsymbol{x}_{i})},
\end{equation}
where $t \in \mathbb{T} = \{0,1\}$ is the ground truth. This formulation is in accordance with prior successful model frameworks such as AlexNet~\cite{AlexNet2012} and VGG-16~\cite{simonyan2014very}, which are also adopted as the base network in some of our reviewed approaches. 

\begin{figure*}[t]
\centering
\includegraphics[width=\linewidth]{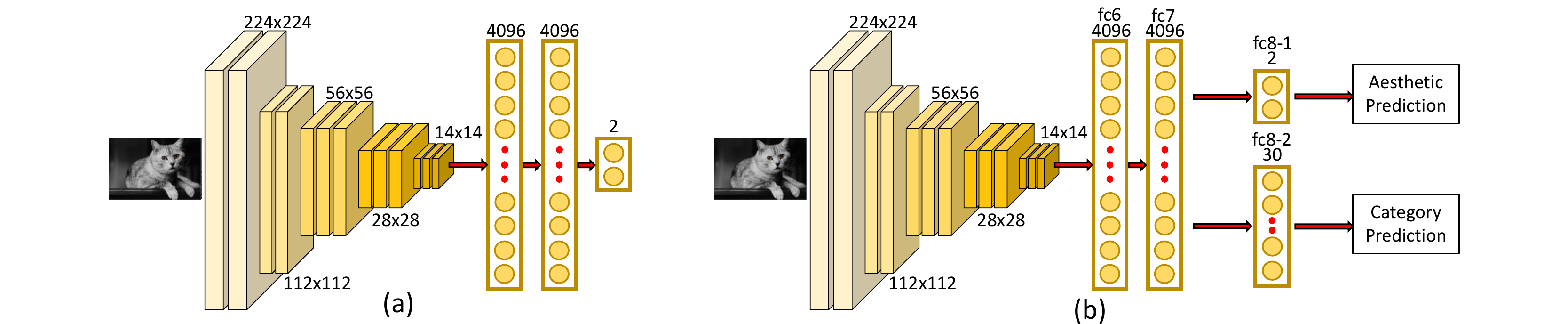}
\vskip -0.45cm
\caption{(a) The structure of the chosen base network for our systematic study on aesthetic quality classification. (b) The structure of the 1-column CNN baseline with multi-task learning.}
\label{fig:1-column-CNN-baseline}
\end{figure*}

The original last fully-connected layer of these two networks are for the  1000-class ImageNet object recognition challenge. For aesthetic quality classification, a 2-class aesthetic classification layer to produce a soft-max predictor is needed (see Fig.~\ref{fig:1-column-CNN-baseline}a). 
Following typical CNN approaches, the input size is fixed to $224 \times 224 \times 3$, which are cropped from a globally warped $256\times256\times3$ images. Standard data augmentation such as mirroring is performed. All baselines are implemented based on the Caffe package~\cite{jia2014caffe}. For clarity of presentation in the following sections, we name the all our fine-tuned baselines as Deep Aesthetic Net (DAN) with the corresponding suffix.

\begin{figure*}[t]
\begin{center}
\includegraphics[width=\linewidth]{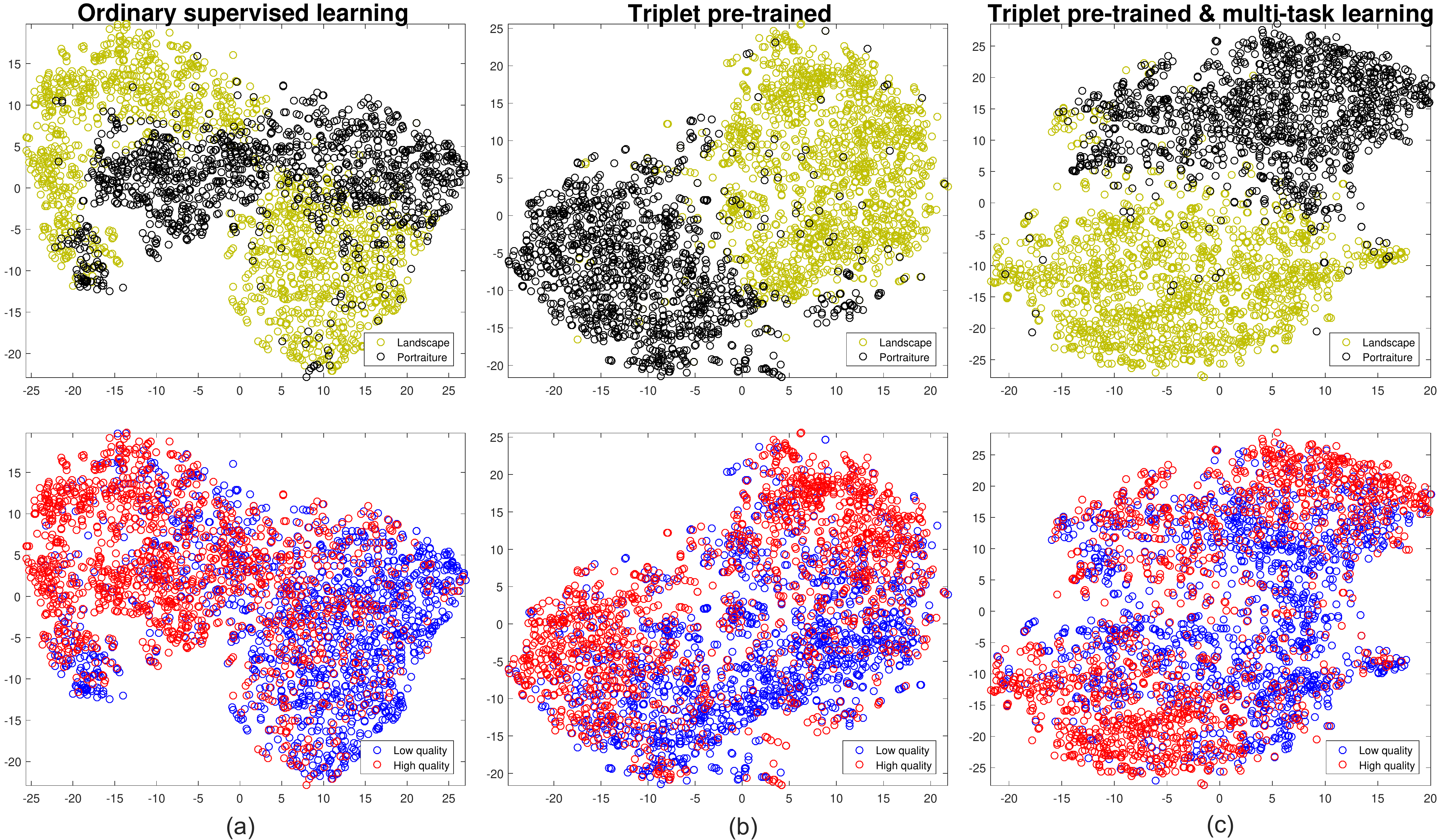}
\vskip -0.45cm
\caption{Aesthetic embeddings of AVA images (testing partition) learned by triplet loss, visualized using t-SNE~\cite{maaten2008visualizing}. (a) ordinary supervised learning without triplet-pretraining and multi-task learning, (b) triplet pre-trained, and (c) combined triplet pre-training and multi-task learning.}
\label{fig:triplet_embedding}
\end{center}
\end{figure*}

\subsection{Training From Scratch vs Fine-Tuning}
Fine-tuning from trained CNN has been proven in~\cite{yosinski2014transferable,dong2015compression} to be an effective initialization approach. The base network of RAPID~\cite{lu2014rapid} uses global image patches and trains a network structure that is similar to AlexNet from scratch. For a fair comparison of similar-depth networks, we first select AlexNet pre-trained with the ILSVRC-2012 training set (1.2 million images) and fine-tune it with the AVA training partition. As shown in Table~\ref{table-scratch-finetune}, fine-tuning from the vanilla AlexNet yields better performance than simply training the base net of RAPID from scratch. Moreover, the DAN model fine-tuned from VGG-16 (see Fig.~\ref{fig:1-column-CNN-baseline}a) yields the best performance in both the balanced accuracy and overall accuracy. It is worth pointing out that other more recent and deeper models such as ResNet~\cite{He2015}, Inception-ResNet~\cite{szegedy2016inception}, or PolyNet~\cite{zhang2017polynet}, could serve as a pre-trained model. Nevertheless, owing to the typically small size of aesthetic datasets, precaution needs be taken during the fine-tuning process. Plausible methods include freezing some earlier layers to prevent overfitting~\cite{yosinski2014transferable}.

\begin{table}[t]
\centering
\caption{Training from scratch v.s. fine-tuning. Using 1-column CNN baseline (\textit{DAN-1}) fine-tuned on AlexNet and VGG-16, both of which are pre-trained on ImageNet dataset.}
\vskip -0.2cm
\label{table-scratch-finetune}
\begin{tabular}{lclc}
\begin{tabular}[c]{@{}c@{}} \\ \textit{Method}  \end{tabular}                         &  \begin{tabular}[c]{@{}c@{}}balanced \\ accuracy  \end{tabular}& & \begin{tabular}[c]{@{}c@{}}overall \\ accuracy\end{tabular}  \\ \hline
RAPID - global~\cite{lu2014rapid}    & -     &   & 67.8   \\
\textit{DAN-1} fine-tuned from AlexNet & 68.0  &   & 71.3   \\
\textit{DAN-1} fine-tuned from VGG-16  & \bf{72.8}    & & 74.1  \\ \hline
\multicolumn{4}{l}{\footnotesize{- The authors in~\cite{lu2014rapid} have not released detailed classification results.}} 
\end{tabular}
\end{table}

\subsection{Mini-Batch Formation}

Mini-batch formation directly affects the gradient direction towards which stochastic gradient descent brings down the training loss in the learning process. We consider two types of mini-batch formation and reveals the impact of this difference on image aesthetic assessment. 

\noindent\textbf{Random sampling}: By randomly selecting examples for mini-batches~\cite{li2014efficient,wilson2003general}, we in fact select from a distribution of the training partition. Since the number of positive examples in the AVA training partition is almost twice as that of the negative examples (Fig.~\ref{fig:teaseImg-AVA}b), models trained with such mini-batches may bias towards predicting positives.

\noindent
\textbf{Balanced formation}: Another approach is to enforce a balanced amount of positives and negatives in each of the mini-batches, \ie for each iteration of backpropagation, the gradient is computed from a balanced amount of positive examples and negative examples. 

Table~\ref{table:DAN-1-column} compares the performance of these two strategies. We observe that although the model fine-tuned with randomly sampled mini-batch reaches a higher overall accuracy, its performance is inferior to the one fine-tuned with balanced mini-batches as evaluated using the balanced accuracy. To keep track of both true positive prediction rates and true negative prediction rates, balanced accuracy is adopted to measure the model robustness on data imbalance issue. Network fine-tune for the rest of the experiments are performed with balanced mini-batches unless otherwise specified.

\begin{table}[t]
\centering
\caption{Effects of mini-batch formation. Using 1-column CNN baseline (\textit{DAN-1}) with VGG-16 as the base network.}
\vskip -0.2cm
\label{table:DAN-1-column}
\begin{tabular}{lclc}
  \begin{tabular}[c]{@{}c@{}}  \\ \textit{Mini-batch formation}  \end{tabular}  & \begin{tabular}[c]{@{}c@{}}balanced \\ accuracy  \end{tabular}& & \begin{tabular}[c]{@{}c@{}}overall \\ accuracy  \end{tabular} \\  \hline
\textit{DAN-1} - Randomly sampled                               & 70.39  &  & 77.65  \\
\textit{DAN-1} - Balanced formation                             & 72.82  &   & 74.06  \\ 
\hline
\end{tabular}
\end{table}

\subsection{Triplet Pre-Training and Multi-Task Learning}
\label{subsec:triplet_multitask}

Apart from directly training using the given training data pairs $\left \{  \boldsymbol{x}_{i}, y_{i} \right \}_{i \in [1, N] }$, one could utilize richer information inherent in the data or auxiliary sources to enhance the learning performance. We discuss two popular approaches below.

\noindent
\textbf{pre-training using triplets}: 
The triplet loss is inspired by Dimensionality Reduction by Learning an Invariant Mapping (DrLIM)~\cite{hadsell2006dimensionality} and Large Margin Nearest Neighbor (LMNN)~\cite{weinberger2005distance}. It is widely used in many recent vision studies~\cite{seguin2016visual,wang2016self,cheng2016person,huang2016learning}, aiming to bring data of the same class closer, while data of different classes further away. 
This loss is particularly suitable to our task - the absolute aesthetic score of an image is arguably subjective but the general relationship that beautiful images are close to each other while images of the opposite class should be apart can hold more easily.

To enforce such a relationship in an aesthetic embedding, one needs to generate mini-batches of triplets, \ie an anchor $\boldsymbol{x}$, a positive instance $\boldsymbol{x}_{+ve}$ of the same class, and a negative instance $\boldsymbol{x}_{-ve}$ of a different class, for deep feature learning. Further, we found it useful to constrain each image triplet to be selected from the same image category.
In addition, we observed better performance by introducing triplet loss in the pre-training stage and continuing with conventional supervised learning on the triplet-pre-trained model.
Table~\ref{table:DAN-1-column-2} shows that the DAN model pre-trained with triplets gives better performance.
We further visualize some categories in the learned aesthetic embedding space in Fig.~\ref{fig:triplet_embedding}. It is interesting to observe that the embedding learned with triplet loss demonstrates much better aesthetic grouping in comparison to that without the use of triplet loss.

\begin{table}[t]
\centering
\caption{Triplets pre-training and multi-task learning. Using 1-column CNN baseline (\textit{DAN-1}) with VGG-16 as the base network. Balanced mini-batch formation is used.}
\vskip -0.2cm
\label{table:DAN-1-column-2}
\begin{tabular}{lclc}
  \begin{tabular}[c]{@{}c@{}}  \\ \textit{Methods}  \end{tabular}  & \begin{tabular}[c]{@{}c@{}}balanced \\ accuracy  \end{tabular}& & \begin{tabular}[c]{@{}c@{}}overall \\ accuracy  \end{tabular} \\  \hline
\textit{DAN-1}                             & 72.82  &   & 74.06  \\ 
\textit{DAN-1} - Triplet pre-trained                                & 73.29 &   & 75.32  \\
\textit{DAN-1} - Multi-task (Aesthetic \& Category)       & 73.39  &  & 75.36  \\
\textit{DAN-1} - Triplet pre-trained + Multi-task       & \bf{73.59} &   & 74.42 \\ \hline
\end{tabular}
\end{table}

\noindent
\textbf{Multi-task learning with image category prediction}:
Can aesthetic prediction be facilitated provided that a model understand to which category the image belongs?  Following the work in~\cite{zhang2016learning} where auxiliary information is used to regularized the learning of the main task, we investigate the potential benefits of using image categories as an auxiliary label in training the aesthetic quality classifier. 

Specifically, given an image labeled with main task label $y$ where $y = 0$ for low-quality image and $y = 1$ for high-quality image, we provide an auxiliary label $c \in C$ denoting one of the image categories, such as ``animals", ``landscape", ``portraits", etc. In total, we include 30 image categories. To learn a classifier for the auxiliary class, a new fully-connected layer is attached to the \textit{fc7} of the vanilla VGG-16 structure to predict a soft-max probability for each of the category classes. The modified 1-column CNN baseline architecture is shown in Fig.~\ref{fig:1-column-CNN-baseline}b. 
\begin{table}[t]
\centering
\caption{Comparison of aesthetic quality classification between our proposed baselines with previous state-of-the-arts on the canonical AVA testing partition.}
\label{AVA-full-table}
\begin{tabular}{lcc}
\begin{tabular}[c]{@{}c@{}}  \\ \textit{Previous work}  \end{tabular}                                                   & \begin{tabular}[c]{@{}c@{}}balanced \\ accuracy\end{tabular} & \begin{tabular}[c]{@{}c@{}}overall \\ accuracy\end{tabular}     \\ \hline
AVA handcrafted features (2012)~\cite{murray2012ava}                                               &  -        & 68.00  \\
SPP (2015)~\cite{lu2015deep}                                               &  -        & 72.85  \\
RAPID - full method (2014)~\cite{lu2014rapid}                                               &  -        & 74.46  \\
Peng et al. (2016)~\cite{peng2016toward}  & - & 74.50 \\
Kao et al. (2016)~\cite{kao2016hierarchical} & - & 74.51\\
RAPID - improved version (2015)~\cite{lu2015improvedrapid}   &   61.77        & 75.42 \\
DMA net (2015)~\cite{lu2015deep}                        &  62.80        & 75.41  \\
Kao et al. (2016)~\cite{kao2016visual} & - & 76.15 \\
Wang et al. (2016)~\cite{wang2016multi} & - & 76.94 \\
Kong et al. (2016)~\cite{kong2016photo} & - & 77.33 \\
Mai et al. (2016)~\cite{mai2016composition} & - & 77.40 \\
BDN (2016)~\cite{wang2016brain}  & 67.99        & 78.08  \\
                                                           &          &        \\
\textit{Proposed baseline using random mini-batches}                                              &          &        \\ \hline
\textit{DAN-1}: VGG-16 (AVA-global-warped-input)                                             & 70.39    & 77.65  \\
\textit{DAN-1}: VGG-16 (AVA-local-patches)                                             & 68.70     & 77.60   \\
Two-column \textit{DAN-2}                       & 69.45    & \bf{78.72} \\ \hline

                                                           &          &        \\
\textit{Proposed baseline using balanced mini-batches}                                              &          &        \\ \hline
\textit{DAN-1}: VGG-16 (AVA-global-warped-input)                                               & \bf{73.59}    & 74.42  \\
\textit{DAN-1}: VGG-16 (AVA-local-patches)                                              				    &  71.40    & 75.8   \\
Two-column \textit{DAN-2}                       		    &  73.51   &  75.96 \\ \hline
\multicolumn{3}{l}{\footnotesize{- The authors in~\cite{murray2012ava,lu2014rapid,lu2015deep,mai2016composition,peng2016toward,kao2016hierarchical,lu2015improvedrapid,kao2016visual,wang2016multi,kong2016photo} have}}    \\
\multicolumn{3}{l}{\footnotesize{\;\; not released detailed results. }}    \\
\end{tabular}
\end{table}
The loss function in Eq.~\eqref{eq:cross_entropy_loss} is now changed to 
\begin{equation}
L_{multiTask} = L(\mathbf{W}) + L_{aux}(\mathbf{W}_{c}),
\end{equation}
\begin{equation}
\begin{multlined}
L_{aux}(\mathbf{W}_{c}) = - \frac{1}{n}\sum_{i =1}^{n}\sum_{c =1}^{C} \{ t_{c}~\text{log}~p(\widehat{y}^{aux}_{c}=t_{c}|\boldsymbol{x}_{i};\mathbf{W}_{c}) \\ \;\;\;\;\;\;\;\;\;\;\;\;\;+ (1-t_{c})\text{log}~p(\widehat{y}^{aux}_{c}=t_{c}|\boldsymbol{x}_{i};\mathbf{W}_{c}))  +\phi(\mathbf{W}_{c}) \},
\end{multlined}
\end{equation}
where $t_{c} \in \{0,1\}$ is the binary label corresponding to each auxiliary class $c \in C$ and $\widehat{y}^{aux}_{c}$ is the auxiliary prediction from the network. Solving the above loss function, the DAN model performance from this multi-task learning strategy is observed to have surpassed the previous one (Table~\ref{table:DAN-1-column-2}). It is worth to note that the category annotation of the AVA-training partition is not complete, with about $25\%$ of images not having categories labeling. For those training instances without categories labels, the auxiliary loss $L_{aux}(\mathbf{W}_{c})$ due to missing labels is ignored. 

\noindent
\textbf{Triplet pre-training + Multi-task learning}:
Combining triplet-pre-training and multi-task learning, the final 1-column CNN baseline reaches a balanced accuracy of $73.59\%$ on the challenging task of aesthetic classification.  The results for different fine-tune strategies is summarized in Table~\ref{table:DAN-1-column-2}.

\noindent
\textbf{Discussion}:
Note that it is non-trivial to boost the overall accuracy at the same time as we try not to overfit the baseline to a certain data distribution. Still, compared with other released results in Table~\ref{AVA-full-table}, with careful training, a 1-column CNN baseline yields strong capability in rejecting false positives while attaining a reasonable overall classification accuracy. We show some qualitative classification results as follows.

\begin{figure*}[t]
\centering
\includegraphics[width=\linewidth]{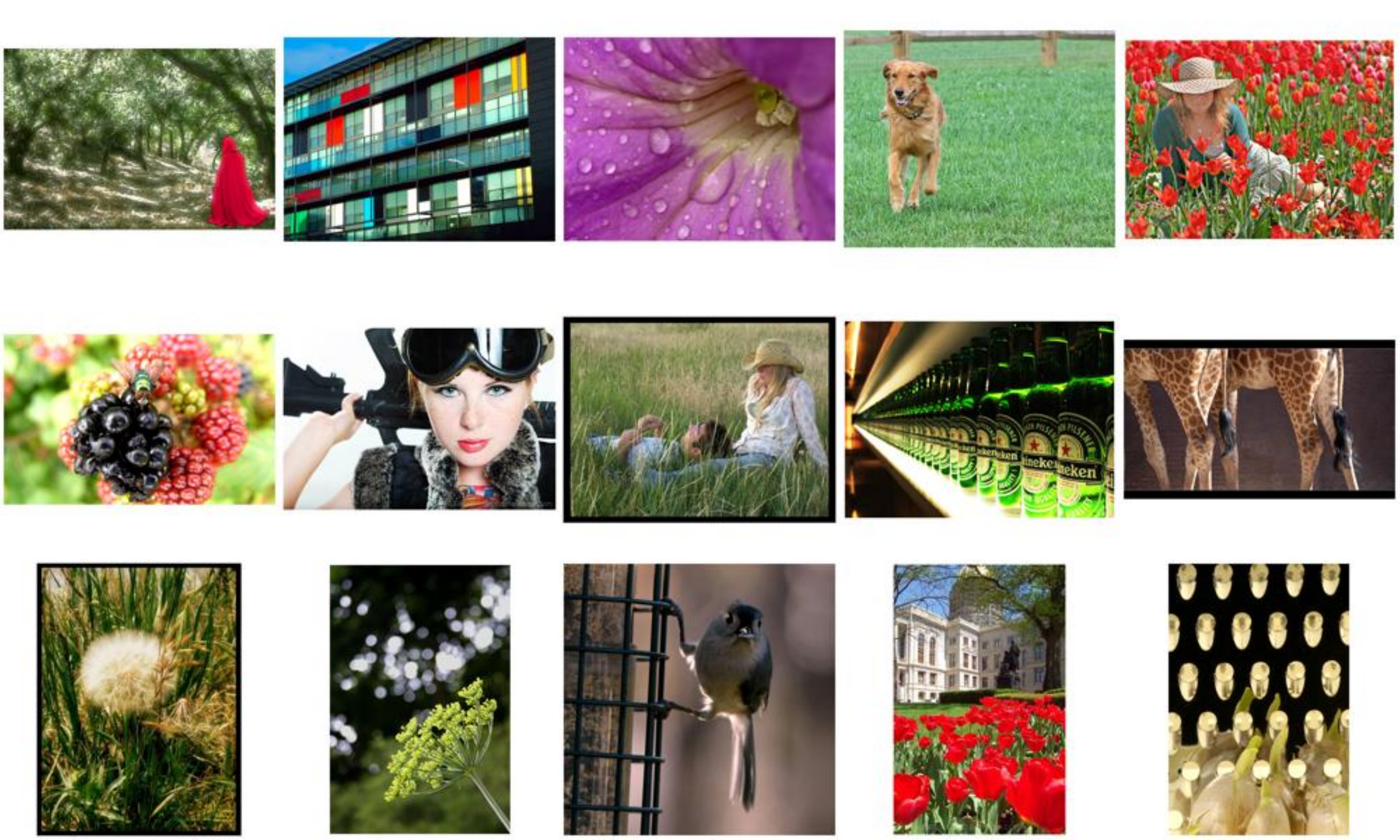}
\caption{Positive examples (high-quality images) that are wrongly classified by BDN and DMA-net but correctly classified by the \textit{DAN-1} baseline.}
\label{fig:positive}
\end{figure*}

\begin{figure*}[!ht]
\centering
\includegraphics[width=\linewidth]{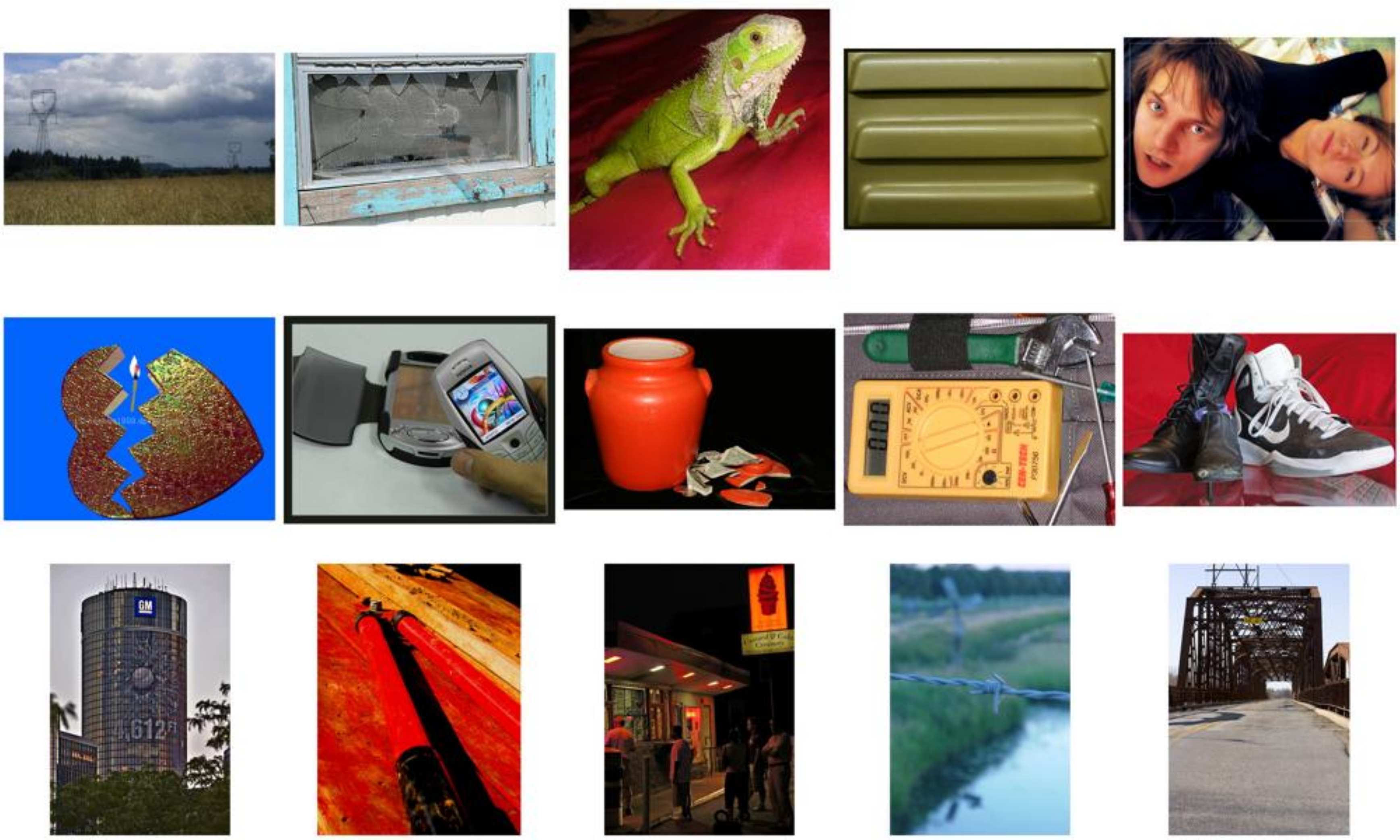}
\caption{Negative examples (low-quality images) that are wrongly classified by BDN and DMA-net but correctly classified by the \textit{DAN-1} baseline.}
\label{fig:negative}
\end{figure*}

\begin{figure*}[t]
\centering
\includegraphics[width=\linewidth]{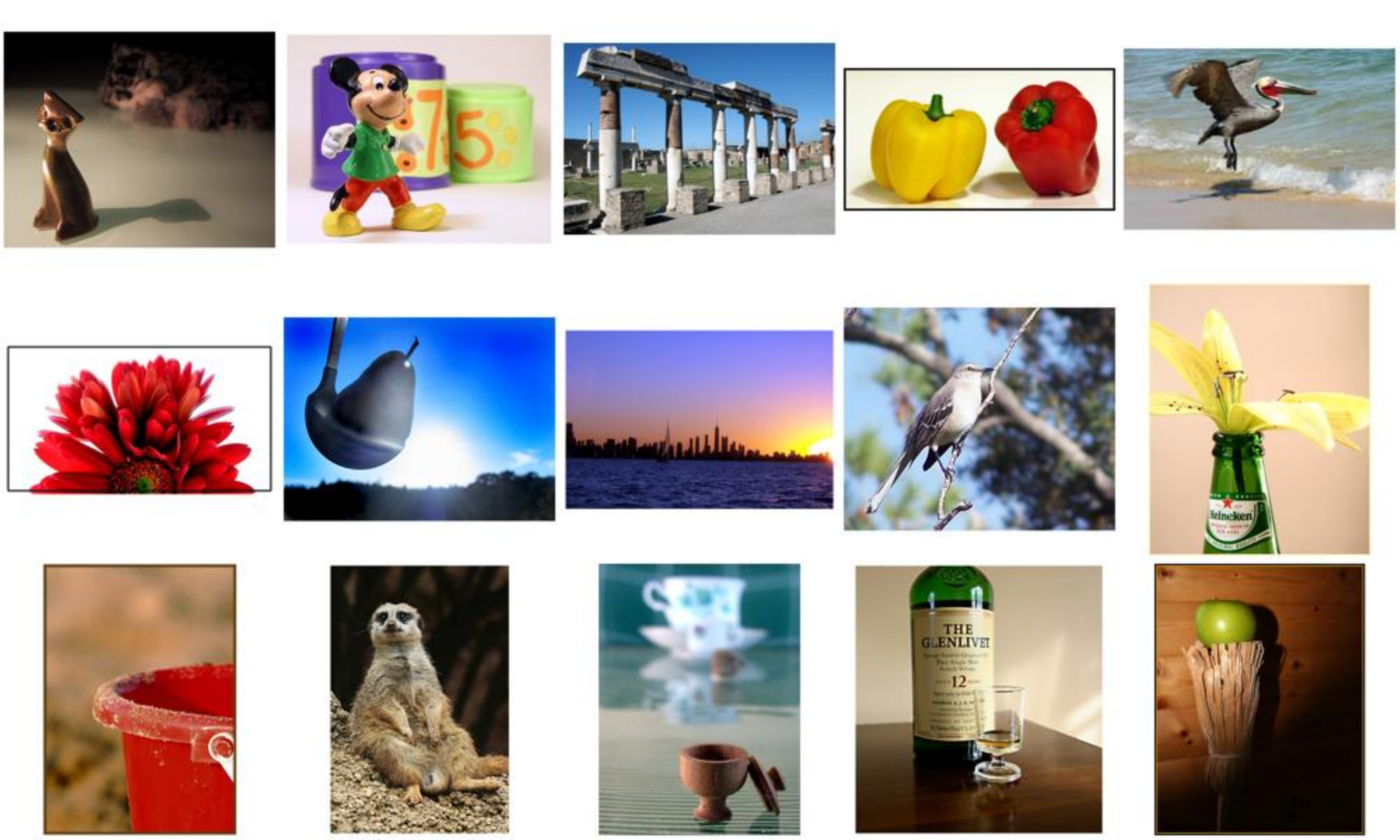}
\caption{Examples with negative groundtruth that get wrongly classified by the \textit{DAN-1} baseline.  High color contrast or depth-of-field is observed in these testing cases.}
\label{fig:DAN_fp}
\end{figure*}

\begin{figure*}[!ht]
\centering
\includegraphics[width=\linewidth]{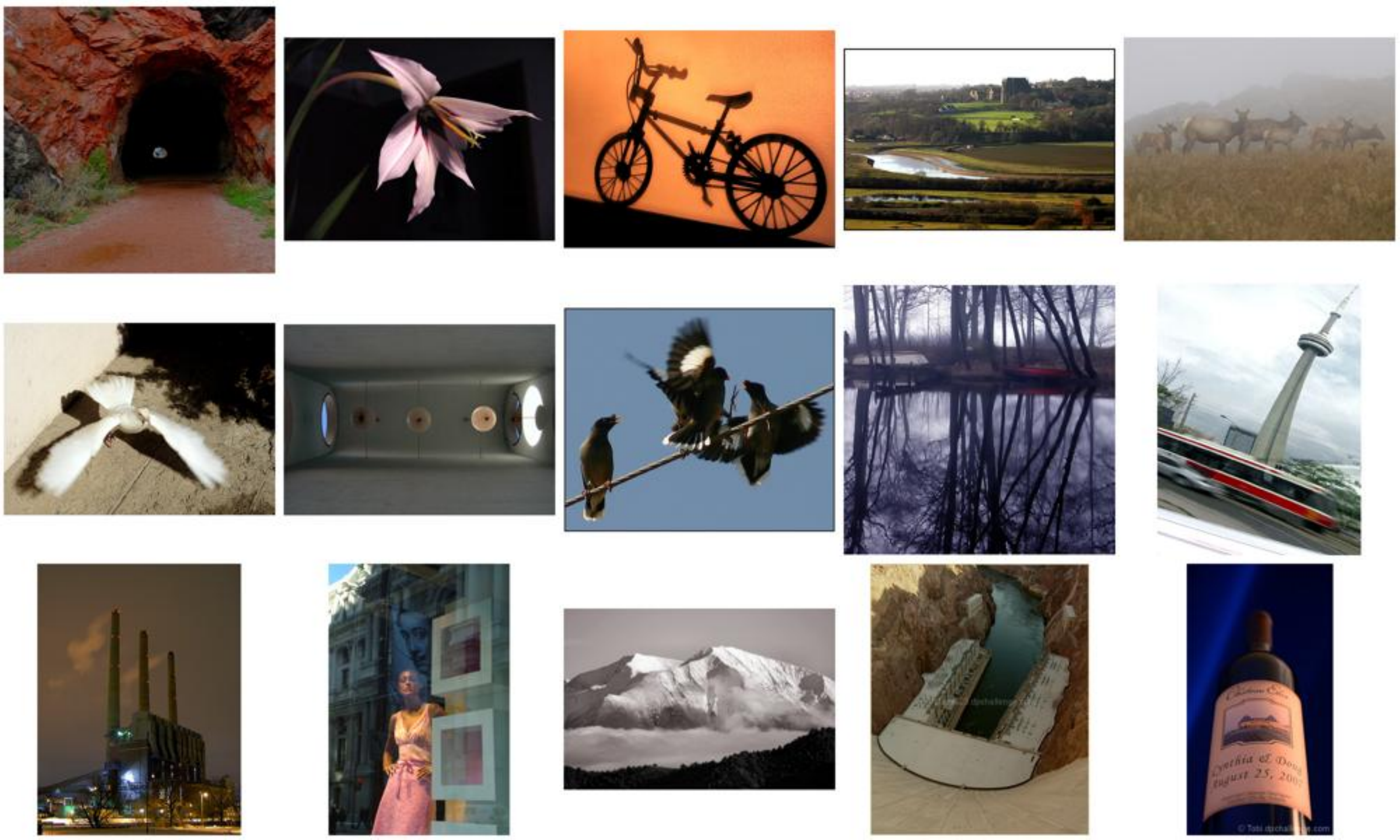}
\caption{Examples with positive groundtruth that get wrongly classified by the \textit{DAN-1} baseline. Most of these images are of low image tones.}
\vskip -0.5cm
\label{fig:DAN_fn}
\end{figure*}

Figures~\ref{fig:positive} and~\ref{fig:negative} show qualitative results of aesthetic classification by the 1-column CNN baseline (using \textit{DAN-1 - Triplet pre-trained + Multi-task}). Note that these examples are neither correctly classified by BDN~\cite{wang2016brain} nor by DMA-net~\cite{lu2015deep}. 
False positive test examples (Fig.~\ref{fig:DAN_fp}) by the \textit{DAN-1} baseline still show a somewhat high-quality image trend with high color contrast or depth-of-field while false negative testing examples (Fig.~\ref{fig:DAN_fn}) mostly reflect low image tones.
Both quantitative and qualitative results suggest the importance of mini-batch formation and fine-tune strategies.

\subsection{Multi-Column Deep Architecture}
\label{subsec:multi-necessary}
\begin{figure}[t]
\centering
\includegraphics[width=\linewidth]{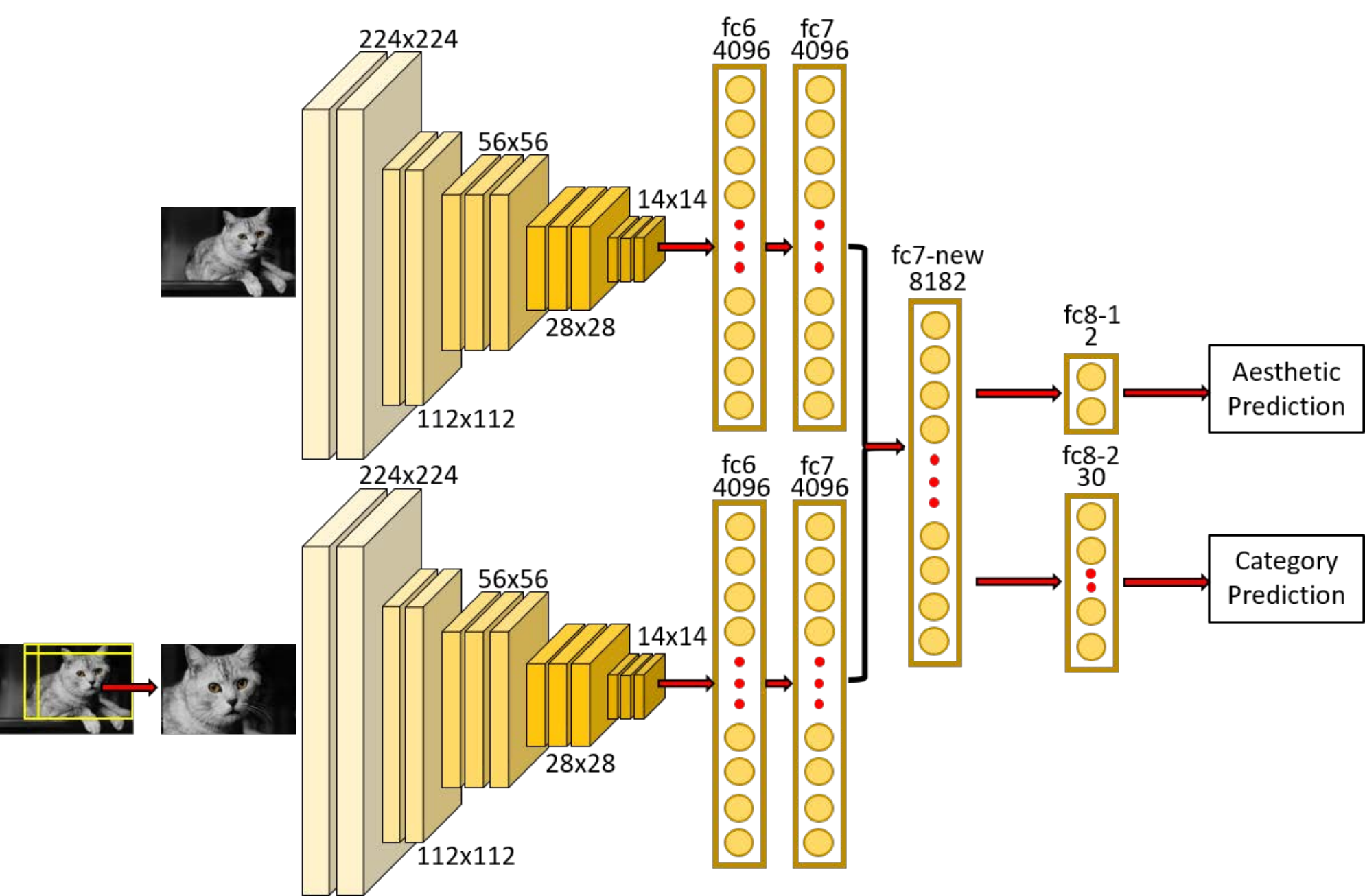}
\vskip -0.3cm
\caption{The structure of the 2-column CNN baseline with multi-task learning.}
\label{fig:2-column-CNN-baseline}
\end{figure}

State-of-the-art approaches~\cite{lu2014rapid,lu2015deep,lu2015improvedrapid,wang2016brain} for image aesthetic classification typically adopt multi-column CNNs (Fig.~\ref{fig:network-double}) to enhance the learning capacity of the model. 
In particular, these approaches
\revision{
benefit from learning multi-scale image information (e.g., global image v.s. local patches) or utilizing image semantic information (e.g., image styles)}. To incorporate insight from previous successful approaches, we prepare another 2-column CNN baseline (\textit{DAN-2}) (see Fig.~\ref{fig:2-column-CNN-baseline}) with focus on the more apparent approach of using local image patches as a parallel input column. 
Both \cite{lu2015deep} and \cite{lu2014rapid} utilize CNNs trained with local image patches as alternative columns in their multi-branch network, with performance evaluated using the overall accuracy. 
For fair comparison, we prepare local image patches of sized $224\times224\times3$ following~\cite{lu2014rapid,lu2015deep} and we fine-tune one \textit{DAN-1} model from the vanilla VGG-16 (ImageNet) with such local patches	. 
Another branch is the original \textit{DAN-1} model, fine-tuned with globally warped input by triplet pre-training and multi-task learning (Sec.~\ref{subsec:triplet_multitask}).
We perform separate experiments where mini-batches of these local image patches are taken from either random sampling or the balanced formation.

As shown in Table~\ref{AVA-full-table}, the \textit{DAN-1} model fine-tuned with local image patches performs inferior under the metric of balanced accuracy compared to the original \textit{DAN-1} model fine-tuned with globally warped input in both random mini-batch learning and balanced mini-batch learning. We conjecture that local patches contain no global and compositional information as compared to globally warped input. Nevertheless, such a drop of accuracy is not observed under the overall accuracy metric.

We next evaluate the 2-column CNN baseline \textit{DAN-2} using the \textit{DAN-1} model fine-tuned with local image patches, and the one fine-tuned with globally warped input. 
We have two variants here depending on whether we employ random or balanced mini-batches.
We observe \textit{DAN-2} trained with random mini-batches attains the highest overall accuracy on the AVA standard testing partition compared to the previous state-of-the-art methods\footnote{Some other works~\cite{tian2015query,kao2015visual,mavridaki2015comprehensive,dong2015multi,dong2015photo} on AVA datasets uses only a small subset of images for evaluation, which is not directly comparable to canonical state-of-the-arts on the AVA standard partition (see Table~\ref{AVA-other-table}).}(see Table~\ref{AVA-full-table}).

Interestingly, we observe that the balanced accuracy of the two variants of \textit{DAN-2} degrades when compared to the respective \textit{DAN-1} trained on globally warped input. 
The observation raises the question if local patches necessarily benefit the performance of image aesthetic assessment. We analyze the cropped local patches more carefully and found that these patches are inherently ambiguous. Thus the model trained with such inputs could easily get biased towards predicting local patch input to be of high-quality, which also explains the performance differences in the two complementary evaluation metrics.

\subsection{Model Depth and Layer-wise Effectiveness}
\label{subsec:equally-finejob}

\begin{figure}[t]
\centering
\includegraphics[width=\linewidth]{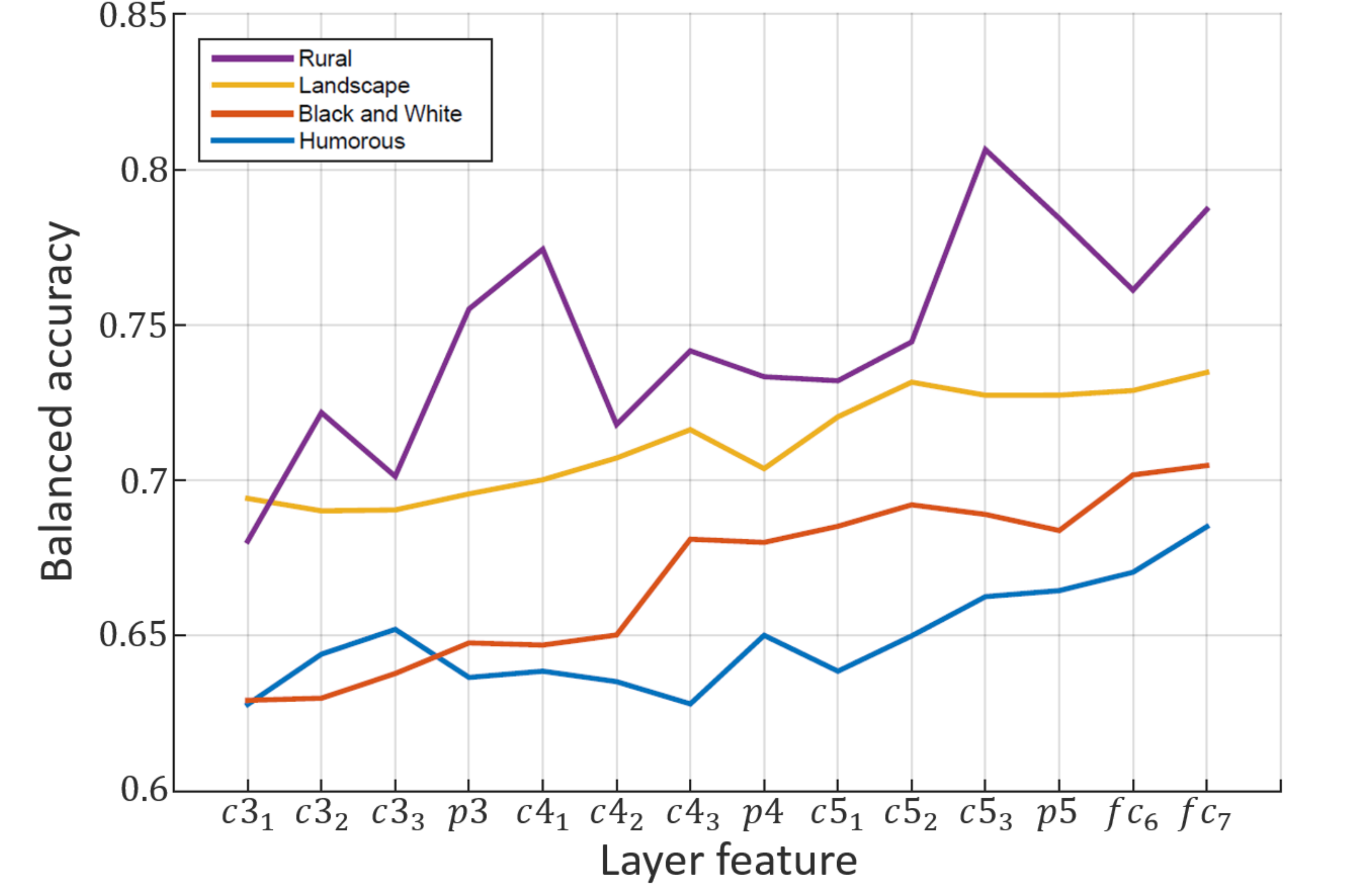}
\vskip -0.3cm
\caption{Layer-by-layer analysis on the difficulties of understanding aesthetics across different categories. From the learned feature hierarchy and the classification results, we observe that image aesthetics in Landscape and Rural categories can be judged reasonably by proposed baselines, yet the more ambiguous Humorous and Black-and-White images are inherently difficult for the model to handle (see also Fig.~\ref{fig:hard-to-learn-1}).
}
\label{fig:hard-to-learn-0}
\end{figure}

Determining the aesthetics of images from different categories takes varying photographic rules. We understand that for some image genre it is not easy to determine its aesthetic quality in general. 
It would be interesting to perform a layer-by-layer analysis and track to what degree a deep model has learned image aesthetics in its hierarchical structure. We conduct this experiment using the 1-column CNN baseline \textit{DAN-1 (Triplet pre-trained + Multi-task)}.
We use layer features generated by this baseline model and train an SVM classifier to perform aesthetic classification on the AVA testing images, and evaluate the performance of different layer features across different image categories. 

Features extracted from convolutional layers of the model are aggregated into a convolutional Fisher representation as done in~\cite{xiehybrid}. Specifically, to extract features from the $d$-th convolutional layer, note that the output feature maps of this $d$-th layer is of size $w \times h \times K$, where $w \times h$ is the size of each of the $K$ output maps. Denote $M^{k}$ as the $k$-th output map. Specifically, a point $M_{i,j}^{k}$ in output map $M^{k}$ is computed from a local patch region \textit{L} of the input image \textit{I} using the forward propagation. By aligning all such points into a vector ${\textbf{\textit{v}}}_{L} = [M_{i,j}^{1}, M_{i,j}^{2}, ..., M_{i,j}^{k}, ..., M_{i,j}^{K}]$, we obtain the feature representation of the local patch region \textit{L}. A dictionary codebook is created using Gaussian Mixture Model from all the $\left \{ \textbf{\textit{v}}_{L} \right \}_{L \in I_{train}}$ and a Fisher Vector representation is subsequently computed using this codebook to describe an input image. The obtained convolutional Fisher representation is used for training SVM classifiers.

\begin{figure*}[t]
\centering
\includegraphics[width=0.825\linewidth]{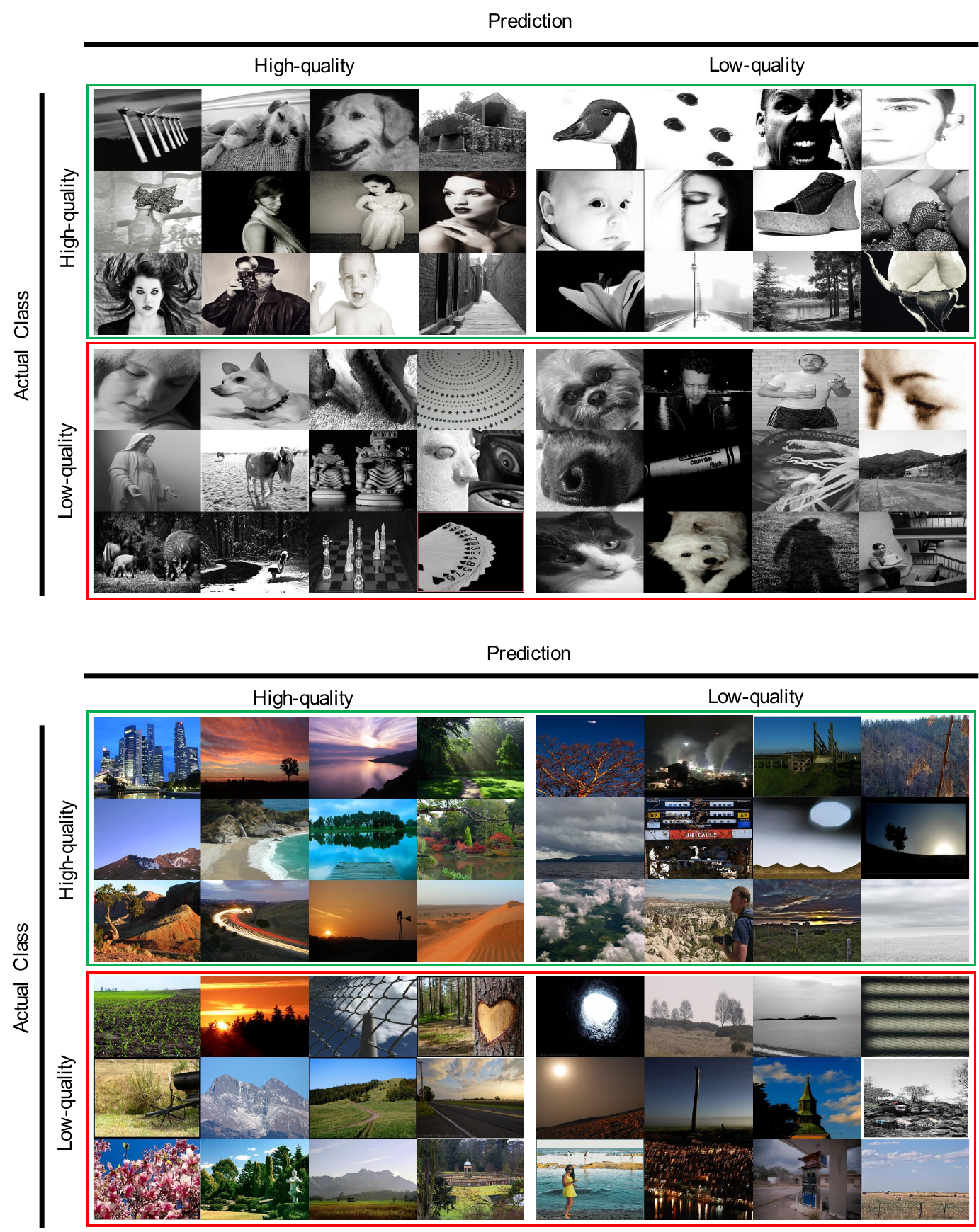}
\vskip -0.3cm
\caption{Layer-by-layer analysis: classification results using the best layer features on \textit{Black-and-White} (top) and \textit{Landscape} (bottom) images.}
\label{fig:hard-to-learn-1}
\end{figure*}

We compare features from layer \textit{conv3\textunderscore 1} to \textit{fc7} of the \textit{DAN-1} baseline and report selected results that we found interesting in Fig.~\ref{fig:hard-to-learn-0}. 
We obtain the following observations: 

\noindent
(1) \textit{Model depth is important} - more abstract aesthetic representation can be learned in deeper layers. The performance of aesthetic assessment can generally be benefited from model depth. This observation aligns with that in general object recognition tasks.

\noindent
(2) \textit{Different categories demand different model depths} - the aesthetic classification accuracy on images belonging to the \textit{Black-and-White} category are generally lower than the accuracy on images of the \textit{Landscape} category across all the layer features. Sample classification results are shown in confusion matrix ordering (see Fig.~\ref{fig:hard-to-learn-1}). High-quality \textit{Black-and-White} images show subtle details that should be considered when assessing their aesthetical level, whereas high-quality \textit{Landscape} images differentiate from those low-quality ones in a more apparent way. 
Similar observations are found, e.g., in \textit{Humorous} and \textit{Rural} categories.
The observation explains why it could be inherently hard for the baseline model to judge whether images from some specific categories are aesthetically pleasing or not, revealing yet another challenge in the assessment of image aesthetics.

\begin{figure*}[t]
\centering
\includegraphics[width=\linewidth]{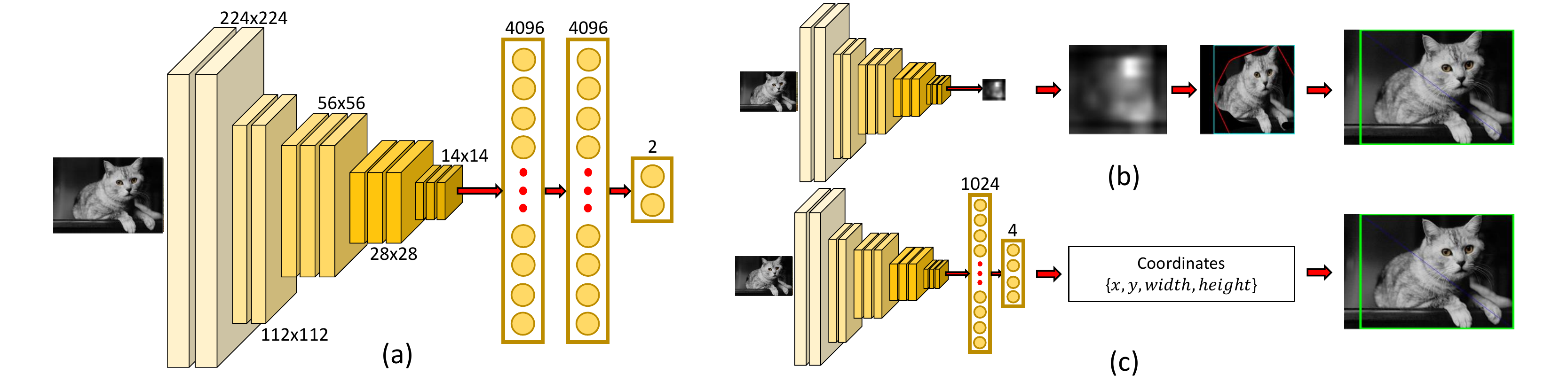}
\vskip -0.3cm
\caption{(a) The originally proposed 1-column CNN baseline. (b) Tweaked CNN by removing all fully-connected layers. (c) Modified CNN to incorporate a crop-regression layer to learn cropping coordinates.}
\label{fig:crop-network}
\end{figure*}


\subsection{From Generic Aesthetics to User-specific Taste}
\label{subsec:user-specific}
\revision{
Individual users may hold different opinions on the aesthetic quality of any single image. One may consider all images presented in Fig.~\ref{fig:DAN_fp} are of high quality to some extent, even though the average scores by the dataset annotators tell otherwise. 
Coping with individual aesthetic bias is a challenging problem. We may follow the idea behind transfer learning~\cite{yosinski2014transferable} and directly model the aesthetic preference of individual users by transferring the learned aesthetic features to fitting personal taste. 
In particular, 
we consider that the \textit{DAN-1} baseline network has already captured a sense of generic aesthetics in the aforementioned learning process; 
to adapt to personal aesthetic preferences, one can include additional data sources for positive training samples that are user-specific, such as the user's personal photographic album or the collection of photos that the user ``liked'' on social media. 
As such, our proposed baseline can be further fine-tuned with ``personal taste'' data for individual users and become a personalized aesthetic classifier. 
}


\section{Image Aesthetic Manipulation}
\label{sec:image_aesthetic_manipulation}
A task closely related to image aesthetic assessment is image aesthetics manipulation, the aim of which is to improve the aesthetic quality of an image. A full review on the techniques of image aesthetics manipulation in the literature is beyond the scope of this survey. Still, we make an attempt to connect image aesthetic assessment to a broader topic surrounding image aesthetics by focusing on one of the major aesthetic enhancement operations, \ie automatic image cropping. 

\subsection{Aesthetic-based Image Cropping}
\label{sec:image_cropping}

\begin{figure*}[t]
\centering
\includegraphics[width=\linewidth]{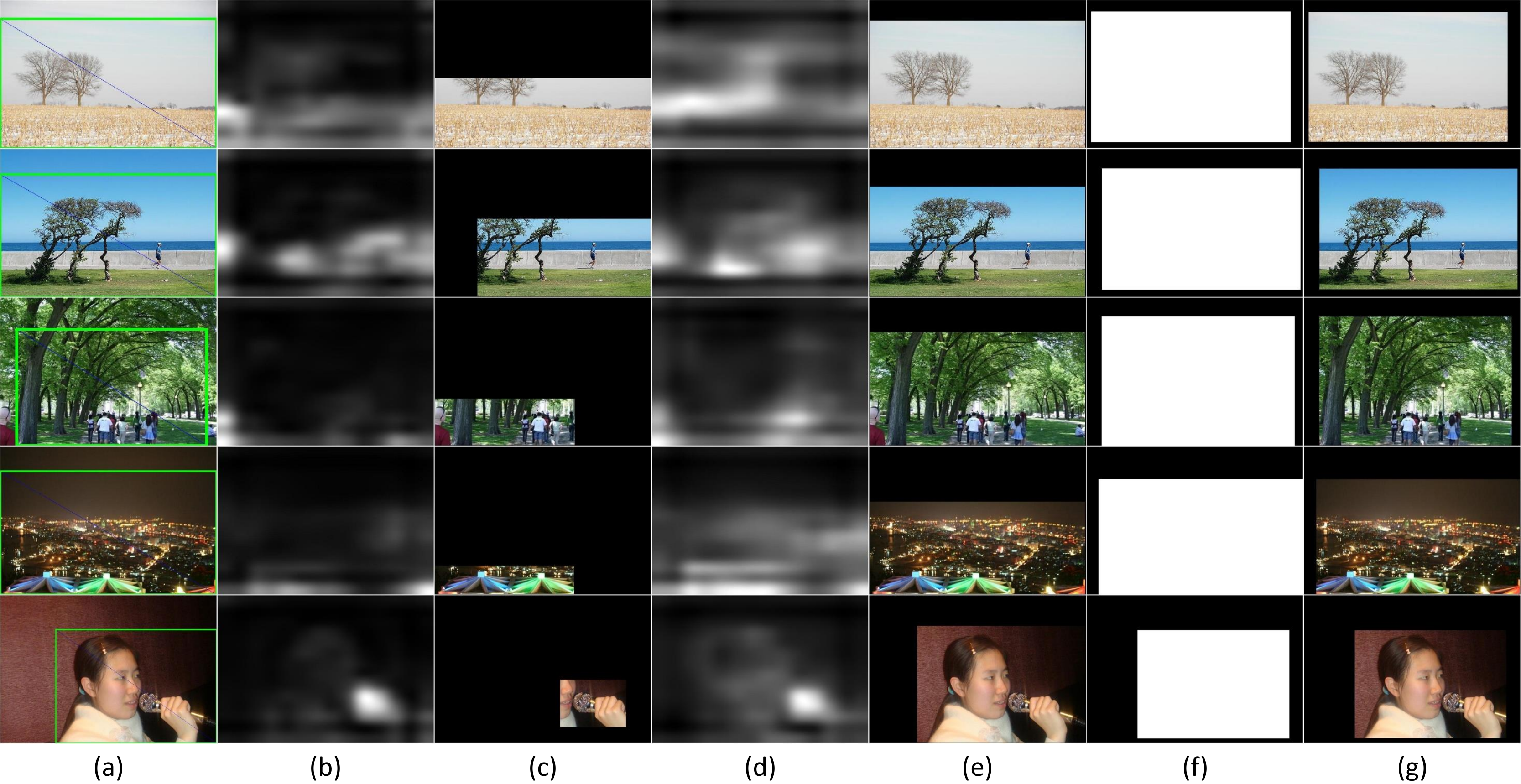}
\vskip -0.3cm
\caption{Layer responses differences of the last conv layer. The images in each row correspond to (a) Input image with ground truth crop; (b) Feature response of the vanilla VGG; (c) Image crops obtained via the feature responses of vanilla VGG; (d) Feature response of the \textit{DAN-1-original} model; (e) Image crops obtained via the \textit{DAN-1-original} model; (f) Four-coordinates window estimated by \textit{DAN-1-regression} network; (g) Cropped image generated by \textit{DAN-1-regression}.}
\label{fig:Crop-responses}
\end{figure*}

Image cropping improves the image aesthetic composition by removing undesired regions from an image, making an image to have higher aesthetic value. A majority of cropping schemes in the literature can be divided into three main approaches. \textbf{Attention/Saliency-based approaches}~\cite{jaiswal2015saliency,sun2013scale,ardizzone2013saliency} typically extract the primary subject region in the scene of interest according to attention scores or saliency maps as the image crops. \textbf{Aesthetics-based approaches}~\cite{nishiyama2009sensation,cheng2010learning,zhang2013probabilistic} assess the attractiveness of some proposed candidate crop windows with low-level image features and rules of photographic composition. 
However, simple handcrafted features are not robust for modeling the huge aesthetic space. 
The state-of-the-art method is the \textbf{Change-based approach} proposed by Yan et al.~\cite{yan2015change,yan2013learning}, which aims at accounting for what is removed and changed by cropping itself and try to incorporate the influence of the starting composition of the initial image on the ending composition of cropped image. This approach produces reasonable crop windows, but the time cost of producing an image crop is prohibitively expensive due to the time spent in evaluating large amounts of crop candidates. 

Automatic thumbnail generation is also closely related to automatic image cropping. Huang el at.~\cite{huang2015automatic} target the visual representativeness and foreground recognizability when cropping and resizing an image to generate its thumbnail. Chen  et al.~\cite{chen2016automatic} aim at extracting the most visually important region as the image crop. Nevertheless, the aesthetics aspects of cropping are not taken into prime consideration in these approaches.

In the next section, we wish to show that high-quality image crops can already be produced from the last convolutional layer of the aesthetic classification CNN. Optionally, this convolutional response can be utilized as the input to a cropping regression layer for learning more precise cropping windows from additional crop data.

\begin{table*}[t]
\centering
\caption{Performance on automatic image cropping. The first number is average overlap ratio, higher is better. The second number (shown in parenthesis) is average boundary displacement error, lower is better.}
\label{cropping-results}
\begin{tabular}{lccccc}
\textit{Previous work}      & *Photographer 1 & &  & Photographer 2 & Photographer 3 \\ \hline
Park et al. \cite{park2012modeling}                             & 0.6034 (0.1062)       & &  &  0.5823 (0.1128) & 0.6085 (0.1102)         \\
Yan et al. \cite{yan2013learning}                              & 0.7487 (0.0667)       & &  &  0.7288 (0.0720) & 0.7322 (0.0719)         \\
Wang et al. \cite{wang2015learning}                             & 0.7823 (0.0623)       & &  &  0.7697 (0.0617) & 0.7725 (0.0701)         \\
Yan et al. \cite{yan2015change}                               & 0.7974 (0.0528)          &  & & $\mathbf{0.7857}$ (0.0567) & 0.7723 (0.0594)  \\
\\
\textit{Proposed baselines}       \\ \hline
Vanilla VGG-16 \ \  (\textit{ImageNet})            & 0.6971 (0.0580)          &  & & 0.6841 (0.0618) & 0.6715 (0.0613)  \\
$\textit{DAN-1-original \ \ \  (\text{AVA training partition})}$            & 0.7637 (0.0437)          &  & & 0.7437 (0.0493) & 0.7360 (0.0495)  \\
$\textit{DAN-1-regression  (\text{cropping data fine-tuned})}$ & \bf{0.8059} (\bf{0.0310})  &  & &  0.7750 (\bf{0.0375}) & \bf{0.7725 (0.0377)   }         \\ \hline
\multicolumn{6}{l}{\footnotesize{$^{*}$There are separate groundtruth annotations by 3 different photographers in the cropping dataset of~\cite{yan2015change}.}}    \\
\multicolumn{6}{l}{}    \\
\end{tabular}
\end{table*}

\subsection{Plausible Formulations based on Deep Models}
\label{subsec:aesthetic_to_crop}
Fine-tuning a CNN model for the task of aesthetic quality classification (Section~\ref{sec:evaluation}) can be considered as a learning process where the fine-tuned model tries to understand the metric of image aesthetics. We hypothesize that the same metric is applicable to the task of automatic image cropping.
We discuss two possible variants as follows. 

\noindent
\textbf{\textit{DAN-1-original} without cropping data} - 
Without utilizing additional image cropping data, a CNN such as the 1-column CNN baseline \textit{DAN-1} can be tweaked to produce image crops
with minor modifications - removing the fully-connected layers. That leaves us with a fully convolutional neural network where the input can be of arbitrary sizes, as shown in Fig.~\ref{fig:crop-network}b. The output of the last convolutional layer of the modified model is of $14 \times 14 \times 512$ dimensional, where the 512 feature maps contain the responses/activations corresponding to the input. To generate the final image crop, we average the 512 feature maps and resize it to the input image size. After that, a binary mask is generated by suppressing the feature map values below a threshold. The output crop window is produced by taking a rectangle convex hull from the largest connected region of this binary mask.

\noindent
\textbf{\textit{DAN-1-regression} with cropping data} - 
Alternatively, to include additional image cropping data $\left \{ \boldsymbol{x}_{i}^{crop}, Y_{i}^{crop} \right \}_{i \in [1, N']} $, where $Y_{i}^{crop} = [x, y, width, height]$,  we follow insights in~\cite{sermanet2013overfeat} and add a window regression layer to learn a mapping from the convolutional response (see Fig~\ref{fig:crop-network}c). As such, we can predict a more precise cropping window by learning this extended regressor from such crop data by a Euclidean loss function:
\begin{align}
L(\mathbf{W}) =  \frac{1}{n}\sum_{i =1}^{n} \left \| \widehat{Y}_{i}^{crop} - Y_{i}^{crop} \right \|^{2}  \;
\end{align}
where $\widehat{Y}_{i}^{crop}$ is the predicted crop window for input image $ \boldsymbol{x}_{i}^{crop}$. 

%
To learn the regression parameters for this additional layer, the image cropping dataset by Yan et al.~\cite{yan2015change} is used for further fine-tune. 
Images in the dataset are labeled with groundtruth crops by professional photographers. Following the evaluation criteria in~\cite{yan2015change}, a 5-fold cross-validation approach is adopted for evaluating the model performance on all images in the dataset. Note that there are only a few hundreds of images in each training fold, hence a direct fine-tune by simply warping the few hundreds of  input to $224 \times 224 \times 3$ could be vulnerable to overfitting. To this end, we fix the weights in the convolutional layers of the \textit{DAN-1-regression} network and only learn the weights for the crop window regression layers. Also, a systematic augmentation approach is adopted as follows. First, input images are randomly jittered for a few pixels ($\times 5$) and mirroring is performed ($\times 2$). Second, we warp the images to have its longer side equal to 224 and hence keeping their aspect ratios. We further downscale the images using a scale of $C \in \left \{50\%, 60\%, 80\%, 90\% \right \}$ ($\times 4$). The downscaled images are then padded back to $224 \times 224$ from $\left \{\text{top-left, top-right, bottom-left, bottom-right} \right \} $ ($\times 4$). Finally, we also have direct input warping regardless of the aspect ratio ($\times 1$). In this manner, one training instance is augmented to $5 \times 2 \times (4 \times 4 + 1) = 170$ input instances. We fine-tune this modified CNN baseline with a learning rate of $10e^{-3}$ and the fine-tuning process converges at around the $2^{nd}$ epoch.

\subsection{Aesthetic-based Image Cropping}
As shown in Fig.~\ref{fig:Crop-responses}, we observe that the convolutional response of the vanilla VGG-16 (ImageNet) for object recognition typically finds a precise focus of the salient object in view, while the 1-column CNN baseline (\textit{DAN-1-original} for aesthetic quality classification) outputs an ``aesthetically-oriented'' salient region where both the object in view and its object composition is revealed. 
Compared to the cropping performance using the vanilla VGG-16, image crops from our \textit{DAN-1-original} baseline already has the capability of removing unwanted regions while preserving the aesthetically salient part in view (see Fig.~\ref{fig:Crop-responses}). The modified CNN (\textit{DAN-1-regression}) further incorporates aesthetic composition information in its crop window regression layer, which serves to refine the crop coordinates for more precise crop generation.

Following the same evaluation settings in~\cite{yan2013learning,yan2015change}, we use average overlap ratio and average boundary displacement error to quantify the performance of automatic image cropping. A higher overlap and a lower displacement between the generated crop and the corresponding groundtruth indicates a more precise crop predictor. As shown in Table~\ref{cropping-results}, directly using the \textit{DAN-1-original} baseline responses to construct image crops already gains competitive cropping performance, while fine-tune it with cropping data (\textit{DAN-1-regression}) further boosts the performance and even surpasses the previous state-of-the-art \cite{yan2015change} on this dataset, especially in terms of boundary displacement error. Last but not least, it is worth to note that CNN-based cropping approach takes merely $\mathtt{\sim}$0.2 seconds for generating an output image crop on GPU and $\mathtt{\sim}$2 seconds on CPU (compared to $\mathtt{\sim}$11 seconds on CPU in~\cite{yan2015change}).

\section{Conclusion and Potential Directions}
Models with competitive performance on image aesthetic assessment have been seen in literature, yet the state of research in this field of study has far from saturated. 
\revision{
Challenging issues 
include the groundtruth ambiguity due to neutral image aesthetics and how to effectively learn category-specific image aesthetics from the limited amount of auxiliary data information.
Image aesthetic assessment can also benefit from an even larger volume of data with richer annotations where every single image is labeled by more users with diverse backgrounds. A large and more diverse dataset will facilitate the learning of future models and potentially allow more meaningful statistics to be captured.}
In this work, we systematically review major attempts on image aesthetic assessment in the literature and further propose an alternative baseline to investigate the challenging problem of understanding image aesthetics. We also discuss an extension of image aesthetic assessment to the application of automatic image cropping  \revision{by adapting the learned aesthetic-classification CNN for the task of aesthetic-based image cropping}. 
We hope that this survey can serve as a comprehensive reference source and 
inspire future research in understanding image aesthetics and 
fostering many potential applications.

\ifCLASSOPTIONcaptionsoff
  \newpage
\fi



%

\bibliographystyle{IEEEtran}
\bibliography{surveyBIB}

\balance

\begin{IEEEbiographynophoto}{Yubin Deng}
received the B.Eng. degree (first-class honors) in Information Engineering from The Chinese University of Hong Kong in 2015. He is currently working
towards the PhD degree at the same department with Hong Kong PhD Fellowship. His research interests include
computer vision, pattern recognition and machine learning. He was a Hong Kong Jockey Club Scholar 2013-2014. He received the champion award of Professor Charles K. Kao Student Creativity Awards in 2015.
\end{IEEEbiographynophoto}

\begin{IEEEbiographynophoto}{Chen Change Loy}
(M'06) received the PhD degree in Computer Science from the Queen Mary University of London in 2010. He is currently a Research Assistant Professor in the Department of Information Engineering, Chinese University of Hong Kong. Previously he was a postdoctoral researcher at Queen Mary University of London and Vision Semantics Ltd. His research interests include computer vision and pattern recognition, with focus on face analysis, deep learning, and visual surveillance. He received the Best Application Paper Honorable Mention at the Asian Conference on Computer Vision (ACCV) 2016. He serves as an Associate Editor of IET Computer Vision Journal and a Guest Editor of Computer Vision and Image Understanding. He is currently a member of IEEE. 
\end{IEEEbiographynophoto}

\begin{IEEEbiographynophoto}{Xiaoou Tang}
(S'93-M'96-SM'02-F'09) received
the BS degree from the University of Science
and Technology of China, Hefei, in 1990, the MS
degree from the University of Rochester, New
York, in 1991, and the PhD degree from the Massachusetts
Institute of Technology, Cambridge,
in 1996. He is a Professor and the Chairman of the Department of Information Engineering, Chinese University of Hong Kong. He worked
as the group manager of the Visual Computing
Group at the Microsoft Research Asia, from 2005 to 2008. His research
interests include computer vision, pattern recognition, and video processing.
He received the Best Paper Award at the IEEE Conference
on Computer Vision and Pattern Recognition (CVPR) 2009. He was a
program chair of the IEEE International Conference on Computer Vision
(ICCV) 2009 and he is an associate editor of the IEEE Transactions on
Pattern Analysis and Machine Intelligence and the International Journal
of Computer Vision. He is a fellow of the IEEE.
\end{IEEEbiographynophoto}

\end{document}